\documentclass[runningheads]{llncs}
\usepackage{calc}
 
\usepackage{eccv}

\usepackage{eccvabbrv}

\usepackage{graphicx}
\usepackage{booktabs}
\usepackage{multirow}
\usepackage{algorithm}
\usepackage{algorithmic}
\usepackage{wrapfig}

\usepackage[accsupp]{axessibility}  

\usepackage{hyperref}

\usepackage{orcidlink}

\usepackage{colortbl}
\usepackage{array}

\definecolor{color3}{rgb}{0.95,0.95,0.95}
\definecolor{color4}{rgb}{0.90,0.9,0.9}
\usepackage{amsfonts}       
\usepackage{graphicx} 
\usepackage{caption} 
\usepackage{float}
\usepackage{xcolor}

\definecolor{HPScolor}{HTML}{74b9ff} 
\definecolor{URcolor}{HTML}{fd79a8}
\definecolor{CLIPcolor}{HTML}{55efc4}

\begin{document}

\title{
From Sparse to Dense: Multi-View GRPO for Flow Models via Augmented Condition Space
}

\titlerunning{Multi-View GRPO}

\author{
Jiazi Bu$^{1,2,6*}$, 
Pengyang Ling$^{3*}$, 
Yujie Zhou$^{1,6*}$, 
Yibin Wang$^{4,9}$, 
Yuhang Zang$^{6}$, \\
Tianyi Wei$^{2}$, 
Xiaohang Zhan$^{7}$, 
Jiaqi Wang$^{9}$, 
Tong Wu$^{8\dag}$, \\
Xingang Pan$^{2\dag}$ \and
Dahua Lin$^{5,6,10}$ \\ 
}  

{
  \renewcommand{\thefootnote}{\fnsymbol{footnote}}
  \footnotetext[1]{Equal contribution. \textsuperscript{\dag}Corresponding author.
  }
}

\authorrunning{J.~Bu et al.}

\institute{
\textsuperscript{\rm 1}Shanghai Jiao Tong University \
\textsuperscript{\rm 2}S-Lab, Nanyang Technological University\\
\textsuperscript{\rm 3}University of Science and Technology of China \
\textsuperscript{\rm 4}Fudan University \\
\textsuperscript{\rm 5}The Chinese University of Hong Kong \
\textsuperscript{\rm 6}Shanghai AI Laboratory 
\textsuperscript{\rm 7}Adobe Research \\
\textsuperscript{\rm 8}Stanford University \
\textsuperscript{\rm 9}Shanghai Innovation Institute \
\textsuperscript{\rm 10}CPII under InnoHK \\
Project Page: \url{https://bujiazi.github.io/mvgrpo.github.io/}
}

\maketitle

\begin{figure}[h]
    \centering
    \vspace{-1em}
    \includegraphics[width=1.0\linewidth]{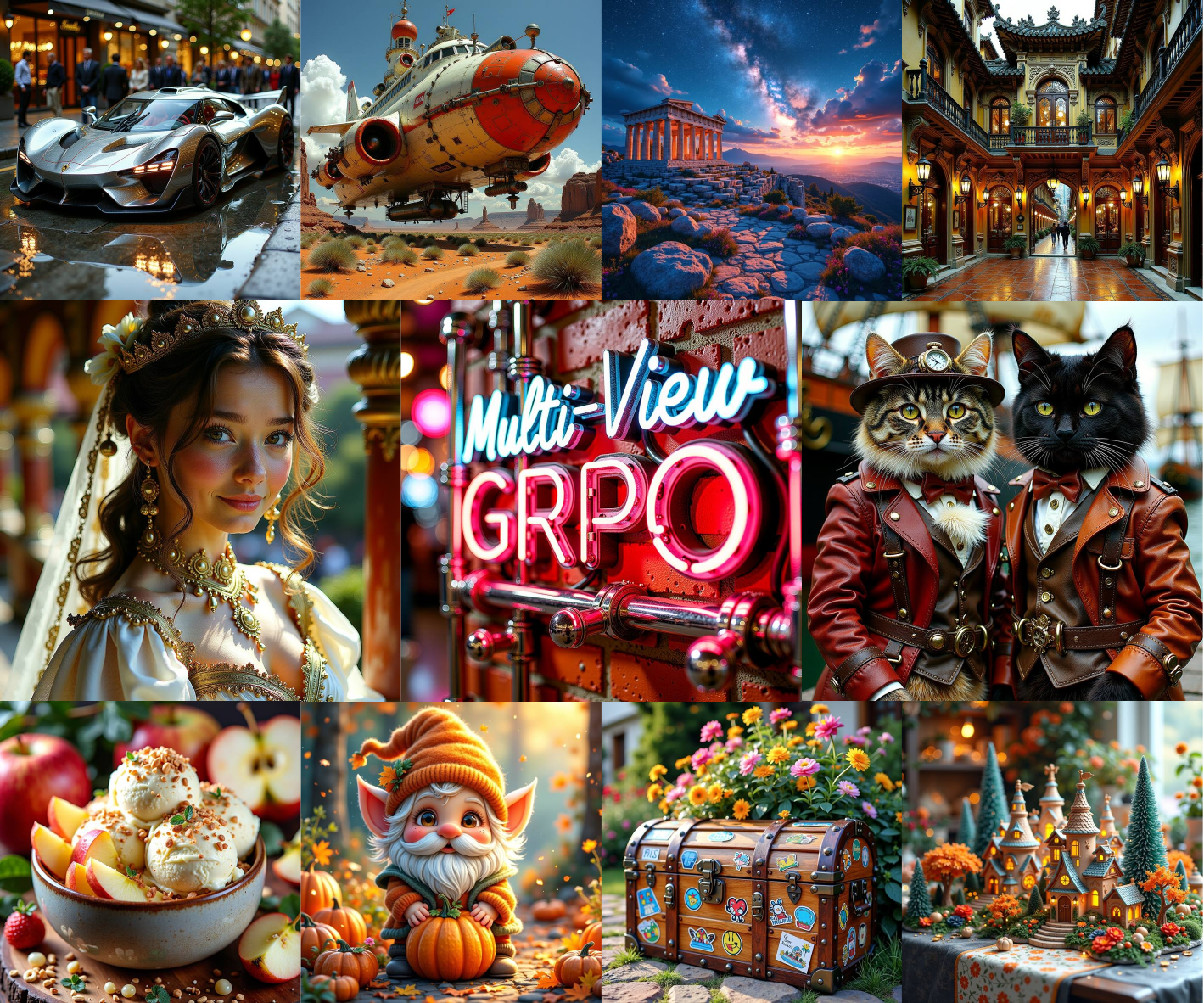}
    \vspace{-2em}
    \caption{
        \textbf{Gallery of MV-GRPO.}
        Our MV-GRPO substantially elevates the generation quality of flow models (Flux.1-dev in this figure), particularly in terms of fine-grained details and photorealism.
        \textbf{Prompts are listed in the supplementary material.}
        }
    \label{fig:teaser}
\end{figure}

\vspace{-2em}
\begin{abstract}
  Group Relative Policy Optimization (GRPO) has emerged as a powerful framework for preference alignment in text-to-image (T2I) flow models. 
  However, we have observed that the standard paradigm that evaluates a group of generated samples against a single condition suffers from insufficient exploration of inter-sample relationships, 
  constraining both alignment efficacy and performance ceilings. 
  To address this sparse single-view evaluation scheme, we propose Multi-View GRPO (\textbf{MV-GRPO}), a novel approach that enhances relationship exploration by augmenting the condition space to create a dense multi-view reward mapping.
  Specifically, for a group of samples generated from one prompt, MV-GRPO leverages a flexible Condition Enhancer to generate semantically adjacent yet diverse captions.
  These captions enable multi-view advantage re-estimation, capturing diverse semantic attributes and providing richer optimization signals. 
  By deriving the probability distribution of the original samples conditioned on these new captions, they can be incorporated into the training process without costly sample regeneration. 
  Extensive experiments demonstrate that MV-GRPO achieves superior alignment performance over state-of-the-art methods.
\end{abstract}

\section{Introduction}
\label{sec:intro}

Over the past few years, diffusion/flow models~\cite{ho2020denoising, song2020denoising,liu2022flow,peebles2023scalable}
have emerged as the dominant paradigm in the generative scheme, 
demonstrating unprecedented capability in synthesizing high-fidelity visual content~\cite{rombach2022high, podell2023sdxl,esser2024scaling,flux2024}.
While pre-training on massive datasets~\cite{schuhmann2022laion,nan2024openvid,chen2024panda} endows these models with impressive generative versatility, ensuring their outputs align with human preferences and task-specific downstream constraints still poses a critical ongoing challenge~\cite{clark2023directly}. Recent advances in Reinforcement Learning (RL)-based post-training paradigms \cite{fan2023dpok, black2023training, rafailov2023direct, schulman2017proximal} have demonstrated considerable efficacy in bridging this gap. 
Through optimization anchored in reward models \cite{wang2025unified, wu2023human, ma2025hpsv3, wang2025unified-think} that faithfully reflect human preferences, these methods effectively align model outputs with desired behaviors and task constraints.

Among these advancements, Group Relative Policy Optimization (GRPO)~\cite{shao2024deepseekmath} has stood out for its efficiency and stability. 
Initially grounded in Large Language Models (LLMs), 
GRPO estimates the advantage of each sample relative to a group average under a given condition (e.g., a textual prompt), thereby eliminating the need for a complex value network and fostering a scalable, flexible framework for preference alignment.
A line of research~\cite{liu2025flow, xue2025dancegrpo,he2025tempflow,Pref-GRPO&UniGenBench} 
have adapted GRPO for visual generation by substituting the standard ODE solvers with SDEs to introduce stochasticity during the flow sampling process.

\begin{figure}[t]
    \centering
    \includegraphics[width=1.0\linewidth]{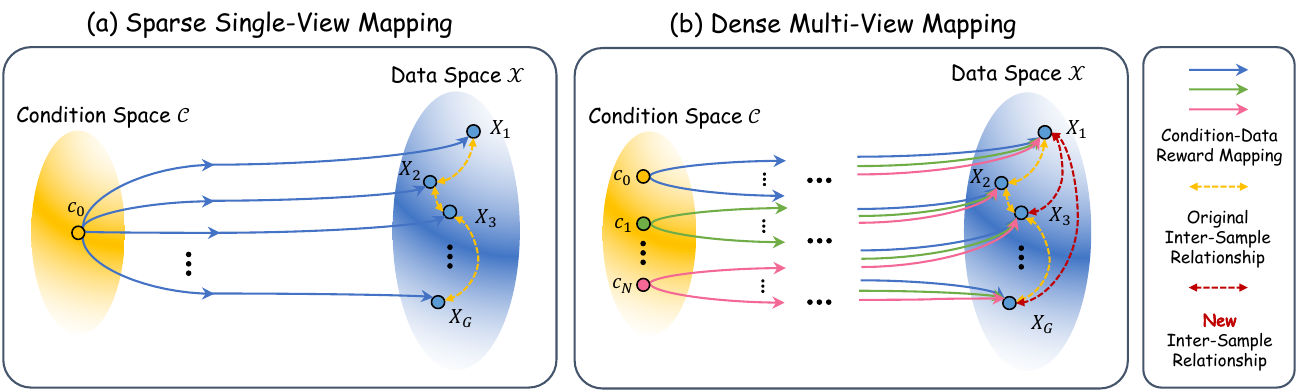}
    \vspace{-2em}
    \caption{
        \textbf{Reward Evaluation in GRPO Training}. (a) Standard flow-based GRPO methods evaluate generated samples under the single original condition, resulting in sparse reward mapping and insufficient inter-sample relationship exploration. (b) Our MV-GRPO leverages an augmented set of conditions to facilitate a dense multi-view mapping, fostering a comprehensive exploration of relationship among samples.
        }
    \label{fig:intro}
    \vspace{-2em}
\end{figure}

As reward estimation relies on noise-free samples generated via computationally expensive iterative denoising, it is essential to fully exploit the relationships among these hard-earned samples for preference alignment.
However, existing methods typically operate under a ``Single-View'' paradigm: they evaluate the generated group solely against the single initial condition. 
This reward evaluation protocol can be reinterpreted as 
\textbf{a sparse, one-to-many mapping from the condition space $\mathcal{C}$ to the data space $\mathcal{X}$}, 
as shown in Fig.~\ref{fig:intro} (a). 
Fundamentally, this paradigm models intra-group relationships by ranking samples based on their alignment with a singular condition, ignoring the multifaceted nature of visual semantics.
For instance, as illustrated in Fig.~\ref{fig:observation}, given a SDE sample describing a cat and a dog within a teacup, it may rank poorly under one condition (``A cat and a dog in a teacup.'') but highly under another similar condition 
specifying visual attributes like lighting, motion or composition.
Consequently, relying solely on the ranking derived from a single prompt is insufficient to gauge the nuanced relationships among samples, resulting in an inherently sparse reward mapping.
In contrast, by incorporating the diverse rankings induced by novel prompts, we can effectively densify the condition-data reward signal. This strategy serves dual purposes: (i) enabling a more comprehensive exploration of intra-group relationships from multiple perspectives, and (ii) establishing intrinsic contrasts by identifying ranking shifts across different conditions, thereby facilitating preference-aligned generation under various conditions.

In light of the above analysis, we propose \textbf{M}ulti-\textbf{V}iew \textbf{GRPO} (\textbf{MV-GRPO}), a novel reinforcement learning framework that provides a dense supervision paradigm via an Augmented Condition Space.  
Specifically, MV-GRPO introduces a flexible \textbf{Condition Enhancer} module to sample a cluster of semantically adjacent descriptors around the original condition anchor.
As depicted in Fig.~\ref{fig:intro} (b), these augmented descriptors, along with the original condition, form a multi-view condition cluster used to jointly evaluate the relative advantage relationships among the generated samples. 
This design offers two key benefits: (i) the multi-view evaluation paradigm reinforces the thoroughness of intra-group sample assessment and inherently facilitates the model's capacity to learn ranking variations under diverse perspectives, promoting heightened awareness of conditional perturbations for enhanced preference alignment, and (ii) by augmenting the condition space $\mathcal{C}$ rather than the computationally expensive data space $\mathcal{X}$, we incur only modest overhead by reusing the hard-earned noise-free samples. Extensive experiments demonstrate that MV-GRPO significantly outperforms standard single-view baselines, achieving superior visual quality and generalization capabilities. Our contribution can be summarized as follows:
\begin{enumerate}
    \item \textbf{Dense Multi-View Mapping}: We identify the sparsity of the single-view reward evaluation in flow-based GRPO and propose a dense, multi-view supervision paradigm via augmenting the condition space.
    \item \textbf{MV-GRPO}: We present MV-GRPO, a novel GRPO framework that leverages a flexible Condition Enhancer to construct an augmented condition set. 
    By re-evaluating the probabilities of original samples under these new conditions, we enable multi-view optimization without costly regeneration.
    \item \textbf{Superior Performance}: MV-GRPO achieves superior performance over existing baselines, excelling in both in-domain and out-of-domain evaluation.
\end{enumerate}
\section{Related Work}

\subsection{Diffusion and Flow Matching}
Diffusion models~\cite{ho2020denoising,song2020denoising,song2020score,dhariwal2021diffusion} have achieved exceptional performance in generative modeling by learning to reverse a gradual noising process, enabling high-fidelity visual synthesis across various modalities~\cite{guo2023animatediff,blattmann2023stable,chen2024videocrafter2,yang2024cogvideox}. 
The introduction of Latent Diffusion Models (LDMs)~\cite{rombach2022high} further reduces the computational cost by performing the diffusion process in a compressed latent space. Instead of simulating a stochastic diffusion path, flow models~\cite{esser2024scaling,lipman2022flow,liu2022flow} directly learn a continuous-time velocity field that moves along straight lines between the noise and data distributions, offering better stability and scalability, and giving rise to numerous state-of-the-art generative models like Flux series~\cite{flux2024, flux-2-2025}, Qwen-Image~\cite{wu2025qwen}, HunyuanVideo series~\cite{kong2024hunyuanvideo,hunyuanvideo2025} and WAN series~\cite{wan2025wan}.

\subsection{Alignment for Diffusion and Flow Models}
Aligning Diffusion and Flow models with human preferences has evolved
from early PPO-style policy gradients~\cite{schulman2017proximal,black2023training,xu2023imagereward} and DPO variants~\cite{rafailov2023direct,wallace2024diffusion,peng2025sudo} toward more 
efficient online reinforcement learning frameworks like Group Relative Policy Optimization (GRPO)~\cite{shao2024deepseekmath}.
To enable GRPO to Flow Matching, foundational works such as Flow-GRPO~\cite{liu2025flow} and DanceGRPO~\cite{xue2025dancegrpo} 
reformulate deterministic Ordinary Differential Equation (ODE) sampling into equivalent Stochastic Differential Equation (SDE) trajectories,
facilitating the stochastic exploration necessary for policy optimization while preserving marginal probability distributions.
Building upon this, 
several variants have emerged to refine the alignment process: 
TempFlow-GRPO~\cite{he2025tempflow} and Granular-GRPO~\cite{zhou2025g2rpo} 
introduce dense credit assignment for precise T2I alignment.
Then, efficiency is further addressed by MixGRPO~\cite{li2025mixgrpo} through a hybrid ODE-SDE sampling 
mechanism and by BranchGRPO~\cite{li2025branchgrpo} via structured branching rollouts. 
DiffusionNFT~\cite{zheng2025diffusionnft} optimizes the forward process directly via flow matching, 
defining an implicit policy direction by contrasting positive and negative generations.
Despite these advancements, 
existing frameworks typically follow a sparse,
one-to-many reward evaluation paradigm,
leading to insufficient and suboptimal exploration. 
In this work, we enable a dense condition-data reward mapping through efficiently augmenting the condition space, achieving more comprehensive advantage estimation and improved alignment performance. 

\section{Method}

\subsection{Preliminary: Flow-based GRPO}

\textbf{Flow Matching as MDP}. 
Flow-based GRPO~\cite{liu2025flow, xue2025dancegrpo} formulates the generation process as a multi-step Markov Decision Process (MDP). 
Let $\mathbf{c} \in \mathcal{C}$ be the condition. The agent $p_\theta$, parameterized by $\theta$, facilitates a reverse-time generation trajectory $\Gamma = (\mathbf{s}_T, \mathbf{a}_T, \dots, \mathbf{s}_0, \mathbf{a}_0)$. Here, the state $\mathbf{s}_t = (\mathbf{c}, t, \boldsymbol{x}_t)$ encompasses the current noisy latent $\boldsymbol{x}_t$ at timestep $t$, initializing from $\boldsymbol{x}_T \sim \mathcal{N}(0, I)$ and terminating at the clean sample $\boldsymbol{x}_0$. The action $\mathbf{a}_t$ corresponds to the single-step denoising update derived from the policy $\pi_\theta(\boldsymbol{x}_{t-1} | \boldsymbol{x}_t, \mathbf{c})$.

\noindent\textbf{Sampling with SDE}. 
Standard flow matching models~\cite{esser2024scaling,flux2024} typically utilize a deterministic Ordinary Differential Equation (ODE) for sampling:
\begin{equation}
    d\boldsymbol{x}_t = \boldsymbol{v}_\theta(\boldsymbol{x}_t, t, \mathbf{c})dt,
    \label{eq:ode}
\end{equation}
where $\boldsymbol{v}_\theta(\boldsymbol{x}_t, t, \mathbf{c})$ is the predicted flow velocity.
To satisfy the stochastic exploration requirements of GRPO,
prior works~\cite{liu2025flow, xue2025dancegrpo} substitute the ODE with a Stochastic Differential Equation (SDE) that preserves the marginal distribution:
\begin{equation}
    d\boldsymbol{x}_t = \left(\boldsymbol{v}_\theta(\boldsymbol{x}_t, t, \mathbf{c}) + \frac{\sigma_t^2}{2t}(\boldsymbol{x}_t + (1-t)\boldsymbol{v}_\theta(\boldsymbol{x}_t, t, \mathbf{c}))\right)dt + \sigma_t d\mathbf{w}_t,
    \label{eq:sde}
\end{equation}
where $d\mathbf{w}_t$ represents the Wiener process increments. The term $\sigma_t = \eta \sqrt{\frac{t}{1-t}}$ modulates the magnitude of injected noise, governed by the hyperparameter $\eta$.
For practical implementation, this is discretized via the Euler-Maruyama scheme:
\begin{equation}
    \boldsymbol{x}_{t+\Delta t} = \boldsymbol{x}_t + \left(\boldsymbol{v}_\theta(\boldsymbol{x}_t, t, \mathbf{c}) + \frac{\sigma_t^2}{2t}(\boldsymbol{x}_t + (1-t)\boldsymbol{v}_\theta(\boldsymbol{x}_t, t, \mathbf{c}))\right)\Delta t + \sigma_t \sqrt{\Delta t} \boldsymbol{\epsilon}, 
    \label{eq:discretization}
\end{equation}
where $\boldsymbol{\epsilon} \sim \mathcal{N}(0, I)$ denotes the Gaussian noise for stochastic exploration.

\noindent\textbf{Training of GRPO}. 
Given a condition $\mathbf{c}$, a generation rollout produces a set of $G$ outputs $\{\boldsymbol{x}_0^i\}_{i=1}^G$. The relative advantage $A_t^i$ of $\boldsymbol{x}_0^i$ is then derived by comparing its reward value $R(\boldsymbol{x}_0^i, \mathbf{c})$ against the aggregate group statistics as follows:
\begin{equation}
    A_t^i = \frac{R(\boldsymbol{x}_0^i, \mathbf{c}) - \text{mean}(\{R(\boldsymbol{x}_0^j, \mathbf{c})\}_{j=1}^G)}{\text{std}(\{R(\boldsymbol{x}_0^j, \mathbf{c})\}_{j=1}^G)}.
    \label{eq:advantage}
\end{equation}
Finally, the policy model is optimized by maximizing the following objective:
\begin{equation}
    \mathcal{J}(\theta) = \mathbb{E}_{\mathbf{c} \sim \mathcal{C}, \{\boldsymbol{x}^i\}_{i=1}^G \sim \pi_{\theta_{\text{old}}}(\cdot|\mathbf{c})} \left[ \frac{1}{G} \sum_{i=1}^G \frac{1}{T} \sum_{t=0}^{T-1} \mathcal{L}_{\text{clip}}(r_t^i, A_t^i) - \beta D_{\text{KL}}(\pi_\theta \| \pi_{\text{ref}}) \right],
    \label{eq:objective}
\end{equation}
where:
\begin{equation}
     \mathcal{L}_{\text{clip}}(r_t^i, A_t^i) = \min \left( r_t^i(\theta)A_t^i, \text{clip}(r_t^i(\theta), 1-\varepsilon, 1+\varepsilon)A_t^i \right),
    \label{eq:loss_function}
\end{equation}
\begin{equation}
r_t^i(\theta) = \frac{p_\theta(\boldsymbol{x}_{t-1}^i | \boldsymbol{x}_t^i, \mathbf{c})}{p_{\theta_{\text{old}}}(\boldsymbol{x}_{t-1}^i | \boldsymbol{x}_t^i, \mathbf{c})}.
\label{eq:ratio}
\end{equation}
The coefficient $\beta$ in Eq.~\ref{eq:objective} balances the KL regularization during training.

\subsection{Observation and Analysis}\label{sec:obs}

\begin{figure}[t]
    \centering
    \includegraphics[width=1.0\linewidth]{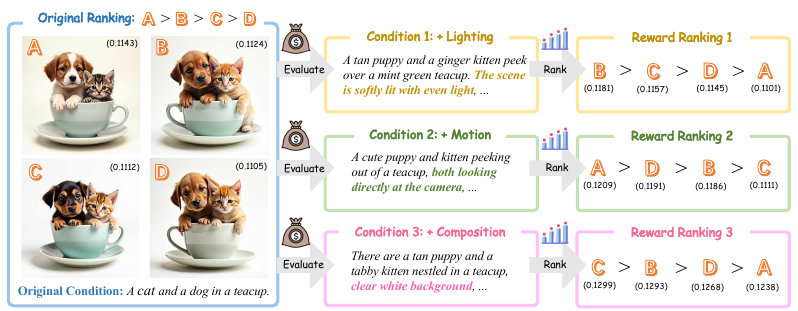}
    \caption{
        \textbf{Reward Ranking Varies with Conditions}. Reward rankings of SDE samples across multiple semantically similar yet different conditions exhibit large variations, indicating that relying on a single condition for advantage estimation is inadequate.
        }
    \label{fig:observation}
\end{figure}

As shown in Fig.~\ref{fig:observation}, given a prompt condition, a set of images can be generated by introducing SDE-based stochasticity into the sampling process. Although these images are consistent with the original prompt in terms of subject content, they also display certain variations, particularly in attributes or local details not specified in the original prompt. Consequently, when evaluating them with the original prompt solely through a single-view paradigm, the influence of such content variations cannot be sufficiently assessed. Notably, when the prompt is perturbed (Condition $1$, $2$ and $3$ in Fig.~\ref{fig:observation}), the relative merits of these images also change accordingly. Intuitively, it is reasonable to perturb the prompt and evaluate the corresponding advantages from the novel perspectives provided by these perturbed prompts, thereby facilitating: (i) a more comprehensive evaluation from diverse viewpoints, and (ii) intrinsic contrastive guidance that teaches the model how advantages shift under different prompt perturbations, thus enhancing its perceptual sensitivity to prompt variations.

\subsection{Condition Enhancer}

To facilitate a comprehensive evaluation of visual samples, we consider sampling auxiliary descriptors from the local manifold surrounding the anchor condition $\mathbf{c}$ in the condition space $\mathcal{C}$ for a dense multi-view assessment. We formalize the Condition Enhancer operator as $\mathcal{E}: \mathcal{C} \times \mathcal{X} \to 2^\mathcal{C}$, which maps an anchor condition $\mathbf{c}$ and a sample group $\mathbf{X}_G = \{\boldsymbol{x}_0^i\}_{i=1}^G$ to an augmented condition set:
\begin{equation}
    \mathcal{V}_K = \mathcal{E}(\mathbf{c}, \mathbf{X}_G) = \left\{ \mathbf{c}' \in \mathcal{C} \mid \mathbf{c}' \sim p_\mathcal{E}(\cdot | \mathbf{c}, \mathbf{X}_G) \right\},
    \label{eq:operator_mapping}
\end{equation}
in which $\mathcal{V}_K$ denotes the resulting augmented condition set containing $K$ additional views, and $p_\mathcal{E}$ represents the sampling distribution of $\mathcal{E}$ given $\mathbf{c}$ and $\mathbf{X}_G$. In practice, we provide two implementations of $\mathcal{E}$:

\noindent\textbf{Online VLM Enhancer}. To dynamically capture the visual semantics of generated samples, a pretrained Vision-Language Model (VLM) is employed as an online Condition Enhancer $\mathcal{E}_\text{VLM}$. During the training loop, $\mathcal{E}_\text{VLM}$ projects each sample $\boldsymbol{x}_0^i \in \mathbf{X}_G$ back to the condition space to obtain posterior descriptors:
\begin{equation}\label{eq:vlm_enhance}
    \mathcal{V}^{\text{post}}_K = \left\{ \mathbf{c}_i^{\text{post}} \in \mathcal{C} \mid \mathbf{c}_i^{\text{post}} \sim p_{\mathcal{E}_\text{VLM}}(\cdot | \mathbf{c}, \boldsymbol{x}_0^i, \texttt{P}_\text{VLM}), i=1 \dots K \right\},
\end{equation}
where the prompt $\texttt{P}_\text{VLM}$ instructs $\mathcal{E}_\text{VLM}$ to describe visual contents within $\boldsymbol{x}_0^i$. For each enhancement given by Eq.~\ref{eq:vlm_enhance}, $\texttt{P}_\text{VLM}$ is randomly sampled from an instruction set $\mathcal{P}_\text{VLM}$ covering diverse descriptive perspectives (e.g., \texttt{lighting}, \texttt{composition}, \texttt{style}, etc.). The above design guarantees the diversity of augmented conditions from two aspects: (i) First, each $\mathbf{c}_i^{\text{post}}$ is derived from a unique SDE sample $\boldsymbol{x}_0^i \in \mathbf{X}_G$; (ii) Second, $\mathcal{E}_\text{VLM}$ is queried with varied instructions $\texttt{P}_\text{VLM} \in \mathcal{P}_\text{VLM}$ focusing on different attributes. In the implementation, we set $K=G$ to fully leverage the generated samples within $\mathbf{X}_G$.

\noindent\textbf{Offline LLM Enhancer}. 
As a complementary strategy based purely on textual semantics,
a pretrained Large Language Model (LLM) is utilized as an offline Condition Enhancer $\mathcal{E}_\text{LLM}$, which directly samples prior descriptors given the anchor condition $\mathbf{c}$:
\begin{equation}\label{eq:llm_enhance}
    \mathcal{V}^{\text{prior}}_K = \left\{ \mathbf{c}_i^{\text{prior}} \in \mathcal{C} \mid \mathbf{c}_i^{\text{prior}} \sim p_{\mathcal{E}_\text{LLM}}(\cdot | \mathbf{c}, \texttt{Mem}, \texttt{P}_\text{LLM}), i=1 \dots K \right\},
\end{equation}
where the prompt $\texttt{P}_\text{LLM}$ instructs $\mathcal{E}_\text{LLM}$ to rewrite the condition. Mirroring the online mode to ensure diversity, (i) $\texttt{Mem}$ represents a historical output buffer introduced to prevent duplicate responses, and (ii) $\texttt{P}_\text{LLM}$ is randomly chosen from an editing prompt set $\mathcal{P}_\text{LLM}$, which includes three operations: \texttt{addition}, \texttt{deletion}, and \texttt{rewriting}. Crucially, since $\mathcal{E}_\text{LLM}$ operates independently of image generation, it can be executed entirely offline before training. 
\textbf{The full details of all VLM and LLM prompts are provided in the supplementary material}.

\begin{figure}[t]
    \centering
    \includegraphics[width=1.0\linewidth]{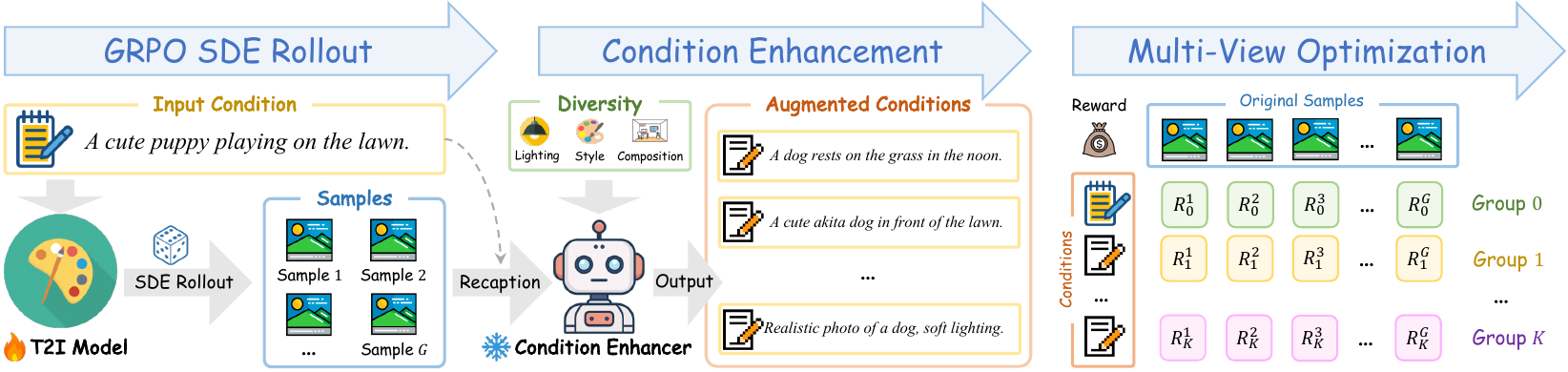}
    \caption{
        \textbf{Overview of MV-GRPO}. 
        MV-GRPO leverages a flexible Condition Enhancer module (a pretrained VLM or LLM) to generate diverse augmented conditions for dense multi-view reward signals, facilitating comprehensive advantage estimation.
        }
    \vspace{-1em}
    \label{fig:pipeline}
\end{figure}

\subsection{Multi-View GRPO}

Building upon the expanded prompts generated through condition enhancement and their associated condition-data mappings, we develop MV-GRPO, a multi-view flow-based GRPO framework that densely couples generated samples with diverse conditions. The overview of MV-GRPO is illustrated in Fig.~\ref{fig:pipeline}.

\begin{figure}[H]
\centering
\resizebox{0.85\linewidth}{!}{ 
\begin{minipage}{\linewidth} 
\begin{algorithm}[H]
\caption{Multi-View GRPO Training Process}
\label{algo:mvgrpo}

\newcommand{\funccommd}[1]{{\textcolor{blue}{#1}}}
\newcommand{\mycommfont}[1]{{\scriptsize\itshape\textcolor{red}{#1}}}

\begin{algorithmic}[1]
    \STATE {\bfseries Require:} Prompt dataset $\mathcal{P}$, policy model $\pi_\theta$, reward model $R$, total sampling steps $T$, SDE sampling timestep set $M$, group size $G$, total training iterations $E$ 
    \STATE {\bfseries Require:} Condition Enhancer $\mathcal{E}$, number of augmented conditions $K$
    \FOR{training iteration $e=1$ {\bfseries to} $E$}
        \STATE Update old policy model: $\pi_{\theta_{\text{old}}} \gets \pi_\theta$
        \STATE Sample batch prompts $\mathcal{P}_b \sim \mathcal{P}$
        \FOR{anchor condition $\mathbf{c} \in \mathcal{P}_b$}
            \STATE \funccommd{\# 1. SDE Sampling for $G$ Samples}
            \STATE Initialize noise: $\boldsymbol{x}_T \sim \mathcal{N}(0,\mathbf{I})$
            \FOR{$t=T$ {\bfseries down to} $1$}
                \IF{$t \in M$}
                    \STATE SDE Sampling: generate $\boldsymbol{x}_{t-1}^i$ for $i=1 \dots G$
                \ELSE
                    \STATE ODE Sampling: generate $\boldsymbol{x}_{t-1}$ 
                \ENDIF
            \ENDFOR
            \STATE Obtain a group of generated samples $\mathbf{X}_G = \{\boldsymbol{x}_0^i\}_{i=1}^G$
            
            \STATE \funccommd{\# 2. Condition Enhancement}
            \STATE Generate augmented condition set $\mathcal{V}_K = \{\mathbf{c}_k\}_{k=1}^K \sim p_{\mathcal{E}}(\cdot | \mathbf{c}, \mathbf{X}_G)$
            
            \STATE \funccommd{\# 3. Multi-View Advantage Estimation}
            \STATE Compute rewards $R(\boldsymbol{x}_0^i, \mathbf{c})$ and advantages $A_t^{i,\mathbf{c}}$ for the anchor condition $\mathbf{c}$
            \FOR{augmented condition $\mathbf{c}_k \in \mathcal{V}_K$}
                \STATE Compute rewards $R(\boldsymbol{x}_0^i, \mathbf{c}_k)$ and advantages $A_t^{i,\mathbf{c}_k}$ 
            \ENDFOR
            
            \STATE \funccommd{\# 4. MV-GRPO Objective Computation}
            \STATE Compute the Multi-View GRPO objective $\mathcal{J}_{\text{MV-GRPO}}(\theta)$ using Eq.~\ref{eq:mv_objective}
        \ENDFOR
        \STATE Update policy: $\theta \gets \theta + \eta\nabla_\theta\mathcal{J}_{\text{MV-GRPO}}(\theta)$
   \ENDFOR
\end{algorithmic}
\end{algorithm}
\end{minipage}
}
\vspace{-2em}
\end{figure}

\noindent\textbf{Training Objective}. The model is fine-tuned on a mixed set of both the original condition and the augmented conditions. The final MV-GRPO objective is constructed by aggregating the policy gradient losses across the anchor view $\mathbf{c}$ and the $K$  augmented conditions in $\mathcal{V}_K$, with the KL term omitted for brevity:
\begin{equation}
\begin{aligned}
    &\mathcal{J}_{\text{MV-GRPO}}(\theta) = \mathbb{E}_{\mathbf{c} \sim \mathcal{C}, \{\boldsymbol{x}_0^i\}_{i=1}^{G} \sim \pi_{\theta_{\text{old}}}(\cdot|\mathbf{c}), \textcolor{blue}{\{\mathbf{c}_k\}_{k=1}^K\sim p_{\mathcal{E}}(\cdot | \mathbf{c}, \mathbf{X}_G)}} \\
    &\Bigg[\underbrace{ \frac{1}{G} \sum_{i=1}^G \frac{1}{T} \sum_{t=0}^{T-1} \min \left( r^i_t(\theta) A_t^{i,\mathbf{c}}, \text{clip}(r^i_t(\theta), 1-\varepsilon, 1+\varepsilon) A_t^{i,\mathbf{c}} \right)}_{\text{Objective for the Original Condition}}
    + \\ & \underbrace{\sum_{k=1}^K \frac{1}{G} \sum_{i=1}^G \frac{1}{T} \sum_{t=0}^{T-1} \min \left( r^{'i}_t(\theta, \mathbf{c}_k) A_t^{i,\mathbf{c}_k}, \text{clip}(r^{'i}_t(\theta, \mathbf{c}_k), 1-\varepsilon, 1+\varepsilon) A_t^{i,\mathbf{c}_k} \right) }_{\text{Objective for Augmented Conditions}}
    \Bigg],
\end{aligned}
\label{eq:mv_objective}
\end{equation}
where $A_t^{i,\mathbf{c}_k}$ is the advantage for the sample $\boldsymbol{x}_0^i$ under an augmented condition $\mathbf{c}_k$ (derived from Eq.~\ref{eq:advantage} by substituting $\mathbf{c}$ with $\mathbf{c}_k$), with $r^i_t(\theta)$ and $r^{'i}_t(\theta, \mathbf{c}_k)$ denoting the importance sampling ratios conditioned on $\mathbf{c}$ and $\mathbf{c}_k$, respectively:
\begin{equation}\label{eq:new_ratio}
    r_t^i(\theta) = \frac{p_\theta(\boldsymbol{x}_{t-1}^i | \boldsymbol{x}_t^i, \mathbf{c})}{p_{\theta_{\text{old}}}(\boldsymbol{x}_{t-1}^i | \boldsymbol{x}_t^i, \mathbf{c})}, r^{'i}_t(\theta, \mathbf{c}_k) = \frac{p_\theta(\boldsymbol{x}_{t-1}^i | \boldsymbol{x}_t^i, \textcolor{blue}{\mathbf{c}_k})}{p_{\theta_{\text{old}}}(\boldsymbol{x}_{t-1}^i | \boldsymbol{x}_t^i, \textcolor{blue}{\mathbf{c}_k})}.
\end{equation}
The training pipeline of MV-GRPO is detailed in Algorithm~\ref{algo:mvgrpo}.

\begin{figure}[t]
    \centering
    \includegraphics[width=1.0\linewidth]{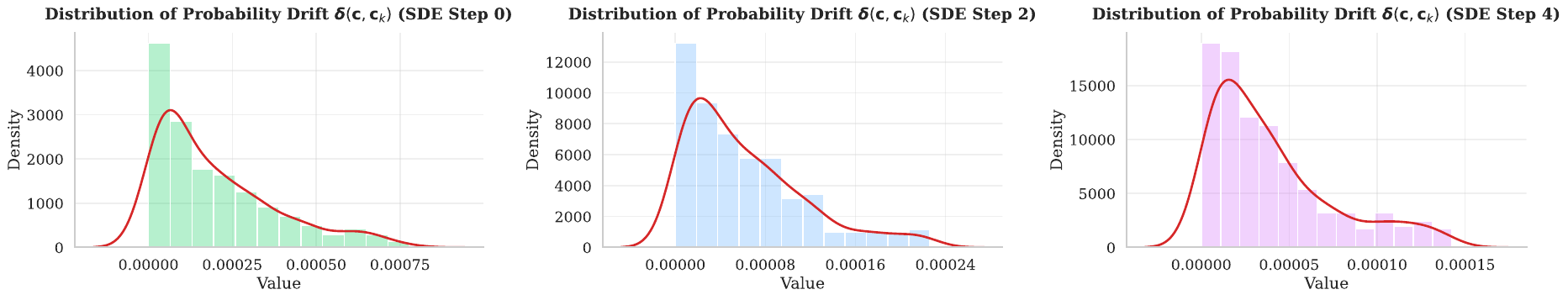}
    \caption{
        \textbf{Distribution of Probability Drift at Different SDE Steps.} Most condition pairs exhibit a drift near zero, demonstrating that the SDE transition probability is effectively preserved 
        when substituting the original with augmented conditions.
        }
    \vspace{-1em}
    \label{fig:distribution}
\end{figure}
\noindent\textbf{Theoretical Perspective}.
To justify optimizing the policy conditioned on an augmented view $\mathbf{c}_k$ using trajectories generated under the anchor $\mathbf{c}$, we examine the transition probability dynamics. 
Recall from Eq.~\ref{eq:discretization} that the single-step transition from $\boldsymbol{x}_t$ to $\boldsymbol{x}_{t-1}$ (denoted as step size $\Delta t$) follows a Gaussian distribution. 
The transition mean $\boldsymbol{\mu}_\theta$ and covariance $\boldsymbol{\Sigma}_t$ derived from the SDE solver are given by:
\begin{equation}
    \boldsymbol{\mu}_\theta(\boldsymbol{x}_t, \mathbf{c}) = \boldsymbol{x}_t + \left(\boldsymbol{v}_\theta(\boldsymbol{x}_t, t, \mathbf{c}) + \frac{\sigma_t^2}{2t}(\boldsymbol{x}_t + (1-t)\boldsymbol{v}_\theta(\boldsymbol{x}_t, t, \mathbf{c}))\right)\Delta t, 
\end{equation}
\begin{equation}
    \boldsymbol{\Sigma}_t = \sigma_t^2 \Delta t \mathbf{I}.
\end{equation}
Consequently, the policy $\pi_\theta(\boldsymbol{x}_{t-1} | \boldsymbol{x}_t, \mathbf{c})$ can be modeled as $\mathcal{N}(\boldsymbol{x}_{t-1}; \boldsymbol{\mu}_\theta(\boldsymbol{x}_t, \mathbf{c}), \boldsymbol{\Sigma}_t)$, where the probability density is formulated as:
\begin{equation}
    p_\theta(\boldsymbol{x}_{t-1} | \boldsymbol{x}_t, \mathbf{c}) = \frac{1}{\sqrt{(2\pi)^d |\boldsymbol{\Sigma}_t|}} \exp \left( -\frac{1}{2} \left\| \boldsymbol{x}_{t-1} - \boldsymbol{\mu}_\theta(\boldsymbol{x}_t, \mathbf{c}) \right\|^2_{\boldsymbol{\Sigma}_t^{-1}} \right).
    \label{eq:prob}
\end{equation}
When evaluating this transition under a new augmented condition $\mathbf{c}_k \in \mathcal{V}_K$, the sampled point $\boldsymbol{x}_{t-1}$ (which was generated via $\mathbf{c}$) is fixed. The probability density of observing this specific transition under the new view $\mathbf{c}_k$ is given by:
\begin{equation}
    p_\theta(\boldsymbol{x}_{t-1} | \boldsymbol{x}_t, \textcolor{blue}{\mathbf{c}_k}) = \frac{1}{\sqrt{(2\pi)^d |\boldsymbol{\Sigma}_t|}} \exp \left( -\frac{1}{2} \left\| \boldsymbol{x}_{t-1} - \boldsymbol{\mu}_\theta(\boldsymbol{x}_t, \textcolor{blue}{\mathbf{c}_k}) \right\|^2_{\boldsymbol{\Sigma}_t^{-1}} \right).
    \label{eq:new_prob}
\end{equation}
The probability drift induced by the condition perturbation is defined as the absolute difference in log-probability densities:
\begin{equation}\label{eq:semantic_drift}
    \boldsymbol{\delta}(\mathbf{c}, \mathbf{c}_k) = \left| \log p_\theta(\boldsymbol{x}_{t-1} | \boldsymbol{x}_t, \mathbf{c}) - \log p_\theta(\boldsymbol{x}_{t-1} | \boldsymbol{x}_t, \mathbf{c}_k) \right|.
\end{equation}
We sampled 500 pairs of $(\mathbf{c}, \mathbf{c}_k)$ through the VLM enhancer and calculate their corresponding probability drift, with the resulting distribution plotted in Fig.~\ref{fig:distribution}. 
Specifically, it can be observed that the drift is minimal for the vast majority of cases across different SDE steps, which is ensured by our Condition Enhancer through sampling semantically adjacent augmented conditions. Given the negligible difference in transition probabilities, $\mathcal{V}_K=\{\mathbf{c}_k\}_{k=1}^K$ offers a meaningful gradient signal for dense supervision and can be seamlessly incorporated into GRPO training. More discussion is provided in the supplementary material. 

\section{Experiments}

\subsection{Implementation Details}

\textbf{Datasets and Models}. Following previous works~\cite{xue2025dancegrpo, li2025mixgrpo, zhou2025g2rpo}, the HPD~\cite{wu2023human} dataset is employed as the prompt dataset. It comprises over 100K prompts for training and a separate set of 400 prompts for evaluation. We adopt Flux.1-dev~\cite{flux2024} as the training backbone, an advanced open-source T2I flow model recognized for its superior visual quality. For the Condition Enhancer, we utilize two leading models from the Qwen series: Qwen3-VL-8B~\cite{bai2025qwen3} is deployed as the online VLM enhancer, 
while Qwen3-8B~\cite{yang2025qwen3} serves as the offline LLM enhancer. Further implementation details are provided in the supplementary material.

\noindent\textbf{Baselines}. The compared methods encompass the vanilla Flux model~\cite{flux2024},
Flow-GRPO~\cite{liu2025flow}, DanceGRPO~\cite{xue2025dancegrpo}, TempFlow-GRPO~\cite{he2025tempflow} and DiffusionNFT~\cite{zheng2025diffusionnft}.

\noindent\textbf{Evaluation Metrics}. To comprehensively assess the effectiveness of MV-GRPO, a diverse set of metrics is employed for evaluation: (i) \textit{Leading VLM-based Reward Models}: HPS-v3~\cite{ma2025hpsv3} and UnifiedReward-v1/v2 (UR-v1/v2)~\cite{wang2025unified};
(ii) \textit{CLIP/BLIP-based Reward Models}: HPS-v2~\cite{wu2023human}, CLIP~\cite{radford2021learning} and ImageReward (IR)~\cite{xu2023imagereward}.

\noindent\textbf{Sampling Details}. Each SDE rollout is conducted with a group size of $G = 12$. The total number of sampling steps is set as $T = 16$ for efficiency. The noise level throughout the sampling process is governed by the hyperparameter $\eta$, which is fixed at $0.7$ in the expression $\sigma_t = \eta \sqrt{\frac{t}{1-t}}$. 
To ensure a fair comparison, all baseline methods adopt the identical configuration described above.

\noindent\textbf{Training Details}. We build MV-GRPO upon Flow-GRPO-Fast~\cite{liu2025flow}, an efficient variant of Flow-GRPO~\cite{liu2025flow}. The training steps are configured as $\{0, 2, 4, 6\}$. Following prior studies~\cite{xue2025dancegrpo, zhou2025g2rpo}, we train MV-GRPO under two experimental settings: (i) \textit{Single-Reward}, where the model is fine-tuned using a single state-of-the-art reward model, specifically either HPS-v3 or UnifiedReward-v2; (ii) \textit{Multi-Reward}, in which HPS-v3 and CLIP are jointly utilized as reward signals to improve training robustness and prevent potential reward-hacking.

\noindent\textbf{Optimization Details}. If not specified, all experiments in this section are conducted on $16\times$NVIDIA H200 GPUs with the batch size setting to $1$. We employ the AdamW optimizer, specifying a learning rate of $2 \times 10^{-6}$ and a weight decay of $1 \times 10^{-4}$. \texttt{bfloat16} (bf16) mixed-precision training is adopted for efficiency.

\subsection{Main Results}

\begin{figure}[t]
    \centering
    \includegraphics[width=1.0\linewidth]{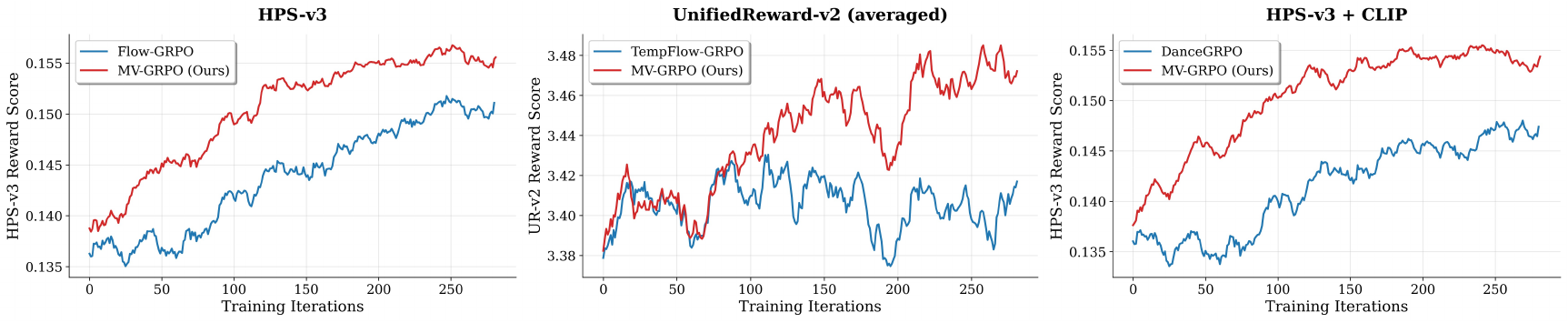}
    \vspace{-2em}
    \caption{
        \textbf{Reward Curves during Training.} Our MV-GRPO outperforms baselines in both convergence speed and performance ceiling under various training settings. 
        }
    \vspace{-1em}
    \label{fig:curve}
\end{figure}
\begin{table}[t]
\centering
\caption{Quantitative comparison of different methods. The best results are in \textbf{bold}, while the
second-best result is \underline{underlined}. UR-v2-A, UR-v2-C, and UR-v2-S denote the Alignment, Coherence, and Style dimensions of UnifiedReward-v2, respectively.}
\vspace{-1em}
\label{tab:results}
\resizebox{\linewidth}{!}{ 
\begin{tabular}{lccccccccc} 
\toprule
\textbf{Reward Model} & \textbf{Method} & {\textbf{HPS-v3}} & {\textbf{UR-v2-A}} & {\textbf{UR-v2-C}} & {\textbf{UR-v2-S}} & \textbf{HPS-v2} & {\textbf{CLIP}} & \textbf{IR} & \textbf{UR-v1} \\ \midrule
/ & Flux.1-dev \cite{flux2024} & 0.133 & 3.250 & 3.634 & 3.223 & 0.305 & 0.389 & 1.042 & 3.604 \\ \midrule
\multirow{6}{*}{{HPS-v3}} 
& Flow-GRPO \cite{liu2025flow} & 0.147 & 3.235 & 3.670 & 3.236 & 0.326 & \underline{0.380} & 1.149 & 3.548 \\
 & DanceGRPO \cite{xue2025dancegrpo} & 0.145 & \underline{3.238} & 3.663 & 3.224 & 0.322 & \textbf{0.381} & 1.165 & \textbf{3.585} \\
 & TempFlow-GRPO \cite{he2025tempflow} & 0.150 & 3.240 & 3.683 & 3.299 & 0.332 & 0.376 & \underline{1.184} & 3.559 \\
 & DiffusionNFT \cite{zheng2025diffusionnft} & 0.149 & \textbf{3.244} & 3.676 & 3.259 & 0.330 & 0.364 & 1.161 & 3.551 \\
 & \cellcolor{color3}{\textbf{MV-GRPO (LLM)}} & \cellcolor{color3}{\underline{0.153}} & \cellcolor{color3}{3.232} & \cellcolor{color3}{\underline{3.692}} & \cellcolor{color3}{\underline{3.313}} & \cellcolor{color3}{\underline{0.336}} & \cellcolor{color3}{0.373} & \cellcolor{color3}{1.176} & \cellcolor{color3}{3.557} \\
 & \cellcolor{color3}{\textbf{MV-GRPO (VLM)}} & \cellcolor{color3}{\textbf{0.155}} & \cellcolor{color3}{3.229} & \cellcolor{color3}{\textbf{3.701}} & \cellcolor{color3}{\textbf{3.320}} & \cellcolor{color3}{\textbf{0.340}} & \cellcolor{color3}{0.370} & \cellcolor{color3}{\textbf{1.193}} & \cellcolor{color3}{\underline{3.569}} \\
 \midrule
\multirow{6}{*}{{UnifiedReward-v2}} 
& Flow-GRPO \cite{liu2025flow} & 0.129 & 3.176 & 3.668 & 3.268 & 0.297 & \underline{0.373} & 1.003 & 3.429 \\
 & DanceGRPO \cite{xue2025dancegrpo} & 0.132 & 3.168 & 3.671 & 3.280 & 0.303 & 0.370 & 1.006 & 3.412 \\
 & TempFlow-GRPO \cite{he2025tempflow} & 0.140 & \underline{3.243} & 3.687 & 3.331 & 0.318 & \textbf{0.374} & \underline{1.140} & 3.433 \\
 & DiffusionNFT \cite{zheng2025diffusionnft} & 0.139 & 3.225 & 3.680 & 3.338 & 0.294 & 0.360 & 1.076 & 3.306 \\
 & \cellcolor{color3}{\textbf{MV-GRPO (LLM)}} & \cellcolor{color3}{\underline{0.142}} & \cellcolor{color3}{3.239} & \cellcolor{color3}{\textbf{3.771}} & \cellcolor{color3}{\underline{3.429}} & \cellcolor{color3}{\underline{0.330}} & \cellcolor{color3}{0.362} & \cellcolor{color3}{1.118} & \cellcolor{color3}{\underline{3.476}} \\
 & \cellcolor{color3}{\textbf{MV-GRPO (VLM)}} & \cellcolor{color3}{\textbf{0.143}} & \cellcolor{color3}{\textbf{3.245}} & \cellcolor{color3}{\underline{3.734}} & \cellcolor{color3}{\textbf{3.454}} & \cellcolor{color3}{\textbf{0.332}} & \cellcolor{color3}{0.367} & \cellcolor{color3}{\textbf{1.187}} & \cellcolor{color3}{\textbf{3.507}} \\
 \midrule
 \multirow{6}{*}{{HPS-v3} + {CLIP}} 
& Flow-GRPO \cite{liu2025flow} & 0.141 & 3.258 & 3.637 & 3.233 & 0.322 & 0.391 & 1.165 & 3.627 \\
 & DanceGRPO \cite{xue2025dancegrpo} & 0.143 & 3.250 & 3.655 & 3.221 & 0.320 & 0.386 & 1.133 & 3.574 \\
 & TempFlow-GRPO \cite{he2025tempflow} & 0.147 & \underline{3.274} & 3.583 & 3.213 & 0.336 & \textbf{0.397} & 1.227 & \underline{3.656} \\
 & DiffusionNFT \cite{zheng2025diffusionnft} & 0.147 & 3.255 & 3.632 & 3.264 & \underline{0.337} & 0.389 & 1.231 & 3.630 \\
 & \cellcolor{color3}{\textbf{MV-GRPO (LLM)}} & \cellcolor{color3}{\underline{0.149}} & \cellcolor{color3}{\textbf{3.276}} & \cellcolor{color3}{\underline{3.695}} & \cellcolor{color3}{\underline{3.347}} & \cellcolor{color3}{\underline{0.337}} & \cellcolor{color3}{0.389} & \cellcolor{color3}{\underline{1.243}} & \cellcolor{color3}{3.650} \\
 & \cellcolor{color3}{\textbf{MV-GRPO (VLM)}} & \cellcolor{color3}{\textbf{0.152}} & \cellcolor{color3}{3.268} & \cellcolor{color3}{\textbf{3.720}} & \cellcolor{color3}{\textbf{3.396}} & \cellcolor{color3}{\textbf{0.342}} & \cellcolor{color3}{\underline{0.393}} & \cellcolor{color3}{\textbf{1.268}} & \cellcolor{color3}{\textbf{3.671}} \\
 \bottomrule
 \vspace{-4em}
\end{tabular}
}
\end{table}

\noindent\textbf{Quantitative Evaluation}.  
As presented in Tab.~\ref{tab:results}, MV-GRPO demonstrates consistent superiority under both single reward (HPS-v3 or UnifiedReward-v2) and multi-reward (HPS-v3 + CLIP) settings.
Specifically, with online VLM condition enhancer, MV-GRPO achieves the best performance across most metrics, particularly excelling in HPS metrics, ImageReward, coherence (UR-v2-C), and style (UR-v2-S), while offline LLM enhancer yields the second-best results. 
This can be attributed to the VLM enhancer's ability to generate tailored sample-specific posterior captions, 
which more precisely describe the generated images and offer more discriminative reward signals 
than the LLM enhancer's prior conditions. 
Furthermore, combining HPS-v3 and CLIP yields notable improvements for both metrics, proving that integrating complementary signals (HPS-v3 for semantic quality, CLIP for text alignment) boosts overall generation.
These results validate our dense multi-view mapping paradigm enables more comprehensive optimization and achieves superior performance. 
The reward curves for the VLM enhancer during training are illustrated in Fig.~\ref{fig:curve}.

\begin{figure}[t]
    \centering
    \includegraphics[width=0.9\linewidth]{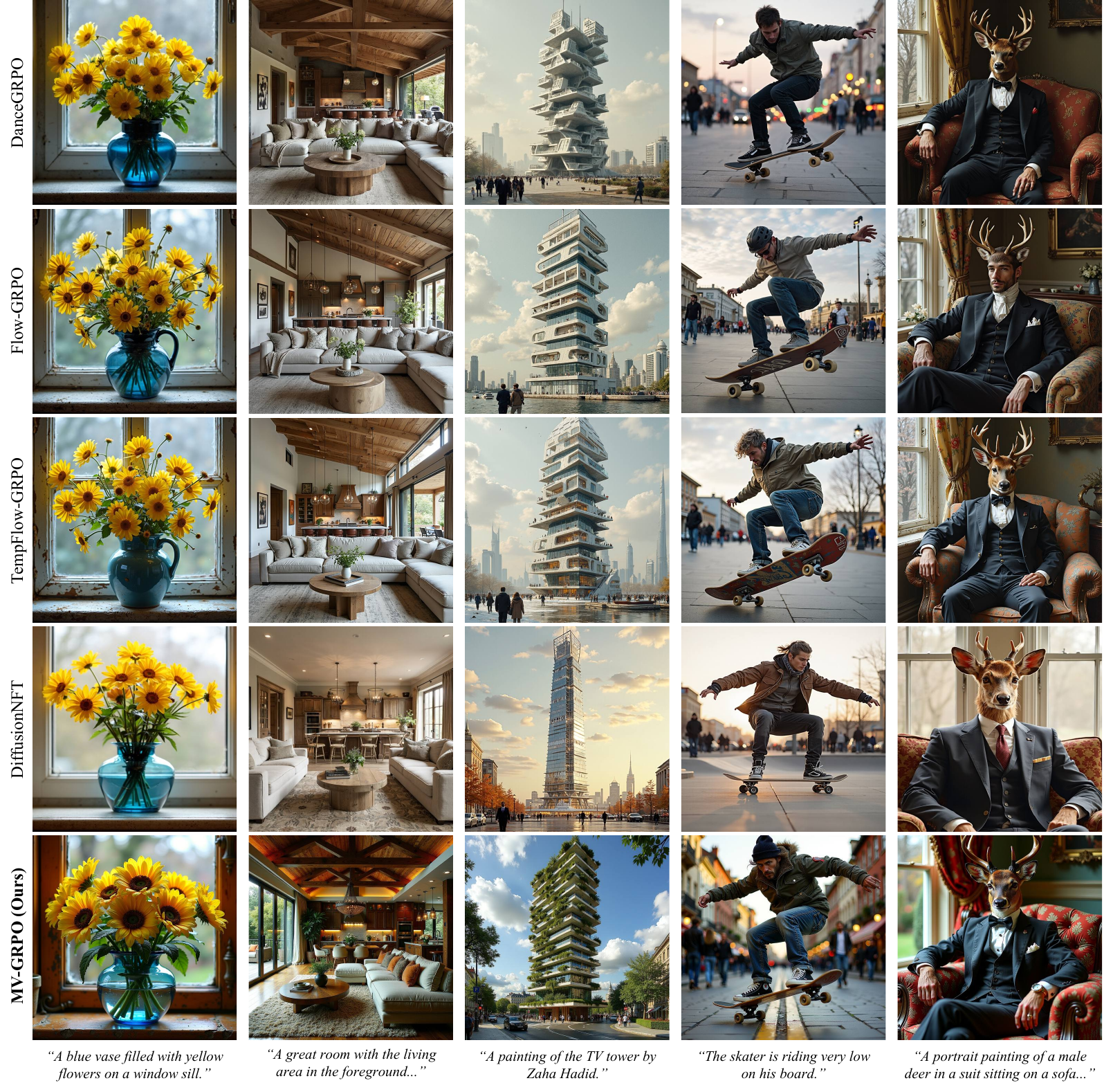}
    \vspace{-1em}
    \caption{
        \textbf{Qualitative Comparisons with Baselines on HPS-v3.}
        }
    \label{fig:hpsv3}
    \vspace{-1.5em}
\end{figure}

\begin{figure}[t]
    \centering
    \includegraphics[width=0.9\linewidth]{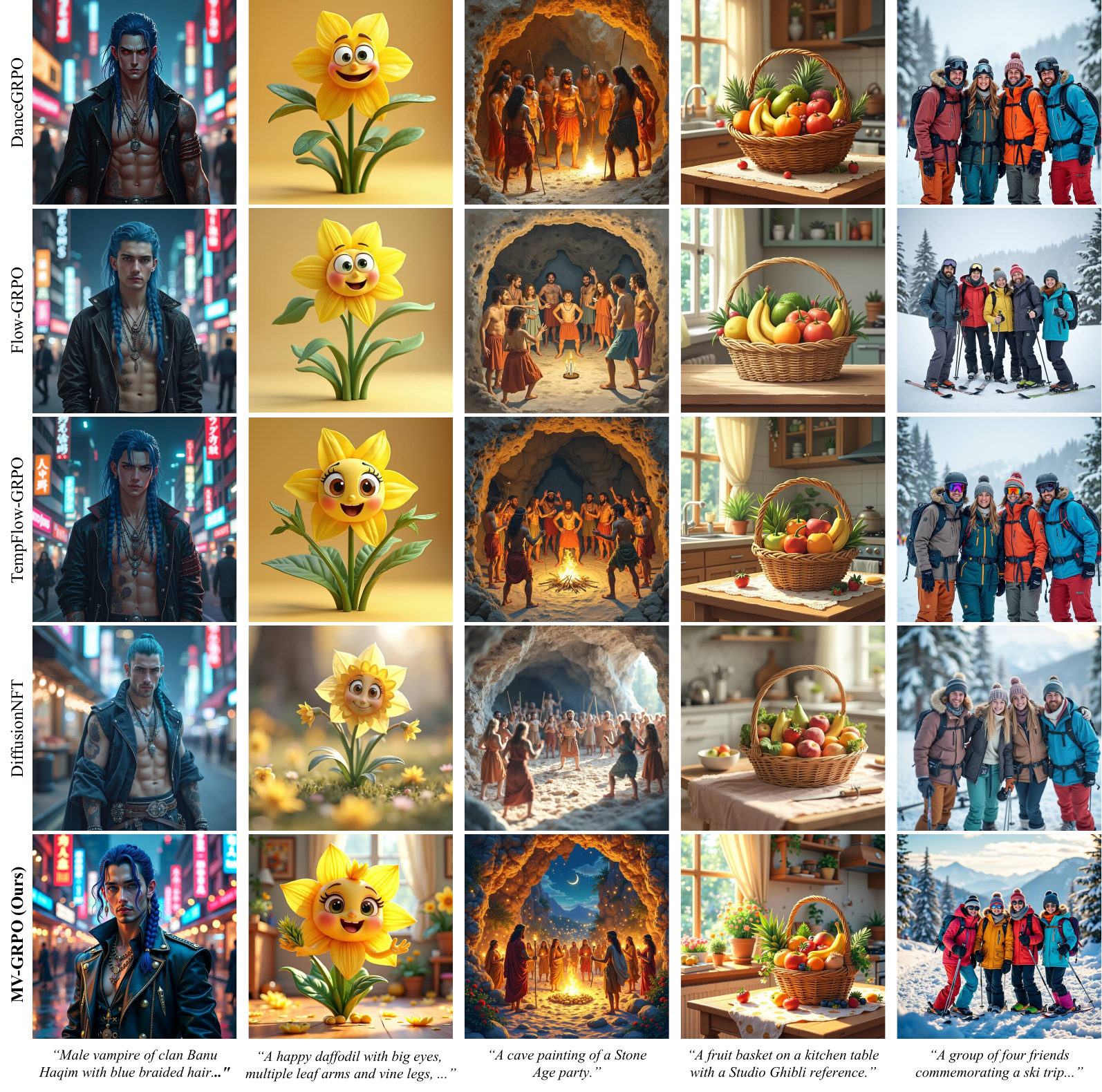}
    \vspace{-1em}
    \caption{
        \textbf{Qualitative Comparisons with Baselines on UnifiedReward-v2.}
        }
    \label{fig:urv2}
    \vspace{-1em}
\end{figure}

\noindent\textbf{Qualitative Comparison}. 
As depicted in Fig.~\ref{fig:hpsv3} and Fig.~\ref{fig:urv2}, 
MV-GRPO consistently outperforms its competitors in semantic alignment, visual fidelity, and structural coherence. 
In the \textit{``room''} and \textit{``tower''} cases (Fig.~\ref{fig:hpsv3}), it renders fine indoor and architectural details with superior clarity.
For the \textit{``skater''} case, MV-GRPO enhances the scene's tension by vividly synthesizing facial expressions and clothing wrinkles. 
Similarly, in the \textit{``daffodil''} and \textit{``cave''} examples (Fig.~\ref{fig:urv2}),
MV-GRPO enriches the compositions with intricate background elements such as furnitures, moons, starry skies, and floral details, 
significantly elevating the cinematic atmosphere and aesthetic appeal of the generated images. 
Finally, in the \textit{``ski''} case, MV-GRPO  not only generates detailed figures but also optimizes the lighting and composition to create a more immersive and expansive snow-covered environment. More results are presented in the supplementary material. 

\begin{table}[H]
    \centering
    \begin{minipage}{0.38\linewidth}
        \centering
        \caption{Comparison in denoiser NFE and iteration time (sec).}
        \vspace{-1em}
        \resizebox{\linewidth}{!}{
        \begin{tabular}{lccc}
            \toprule
            \textbf{Method} & \textbf{NFE} & \textbf{Iteration Time ($s$)} \\
            \midrule
            Flow-GRPO-Fast~\cite{liu2025flow} & \textbf{13} & \textbf{156.26} \\
            + Data Augmentation &  \underline{156} & 1931.15 \\
            \cellcolor{color3}{\textbf{+ MV-GRPO (Ours)}} & \cellcolor{color3}{\textbf{13}} &  \cellcolor{color3}{\underline{191.95}} \\
            \bottomrule
        \end{tabular}}
        \label{tab:latency}
    \end{minipage}
    \hfill
    \begin{minipage}{0.60\linewidth}
    \centering
    \caption{Compatibility with Other GRPO Frameworks. HPS-v3 is employed as the reward model.}
    \vspace{-1em}
    \resizebox{\linewidth}{!}{
    \begin{tabular}{lcccccc}
        \toprule
        \textbf{Method}  & \textbf{HPS-v3} & \textbf{UR-v2 (avg)} & \textbf{HPS-v2} & \textbf{CLIP} & \textbf{IR} & \textbf{UR-v1} \\
        \midrule
        DanceGRPO-Fast~\cite{xue2025dancegrpo} & 0.149 & 3.387 & 0.330 & \textbf{0.375} & 1.108 & 3.555 \\
        \cellcolor{color3}{\textbf{+ MV-GRPO (LLM)}} & \cellcolor{color3}{\underline{0.153}} & \cellcolor{color3}{\underline{3.411}} & \cellcolor{color3}{\underline{0.334}} & \cellcolor{color3}{\underline{0.370}} & \cellcolor{color3}{\underline{1.172}} & \cellcolor{color3}{\underline{3.561}} \\  
        \cellcolor{color3}{\textbf{+ MV-GRPO (VLM)}} &
        \cellcolor{color3} \textbf{0.154} &
        \cellcolor{color3} \textbf{3.418} &
        \cellcolor{color3} \textbf{0.342} &
        \cellcolor{color3} 0.368 &
        \cellcolor{color3} \textbf{1.187} &
        \cellcolor{color3} \textbf{3.567} \\
        \bottomrule
    \end{tabular} }
    \label{tab:dance}
\end{minipage}
\end{table}
\vspace{-3em}
\subsection{Additional Analysis}

\noindent\textbf{Comparison in Latency}. As shown in Tab.~\ref{tab:latency}, our MV-GRPO introduces only a modest overhead compared with its baseline method, nearly $10\times$ less than that of applying an equal amount of data augmentation. Moreover, since MV-GRPO requires no sample regeneration, it matches the baseline in terms of the denoiser NFE, further demonstrating the efficiency of augmenting the condition space.

\noindent\textbf{Compatibility with Other GRPO Frameworks}. Similar to Flow-GRPO~\cite{liu2025flow}, DanceGRPO~\cite{xue2025dancegrpo} stands as a foundational work in flow-based GRPO.
As depicted in Tab.~\ref{tab:dance}, we integrate MV-GRPO into DanceGRPO-Fast (DanceGRPO equipped with the few-step training of Flow-GRPO-Fast) and achieve remarkable performance improvements, highlighting its flexibility and versatility.

\subsection{Ablation Study}

We conducted ablation studies on the online VLM enhancer setting trained with reward model HPS-v3. The results are presented in Tab.~\ref{tab:ablation}.

\noindent\textbf{Effects of Condition Number}. The number of augmented conditions $K$ determines the density of the condition-data reward mapping. Empirically, a larger $K$ facilitates a more thorough exploration of intra-group relationships, leading to better optimization and superior performance.
As can be observed in Tab.~\ref{tab:ablation} (a), performance improves with $K$ but tends to saturate at higher values.
Notably, even with $K = G/2 = 6$, the model achieves competitive performance with $K = G = 12$. However, given that the overhead of condition augmentation is small, $K = G = 12$ is chosen to maximize the density of the reward mapping.

\begin{figure}[t]
    \centering
    \includegraphics[width=1.0\linewidth]{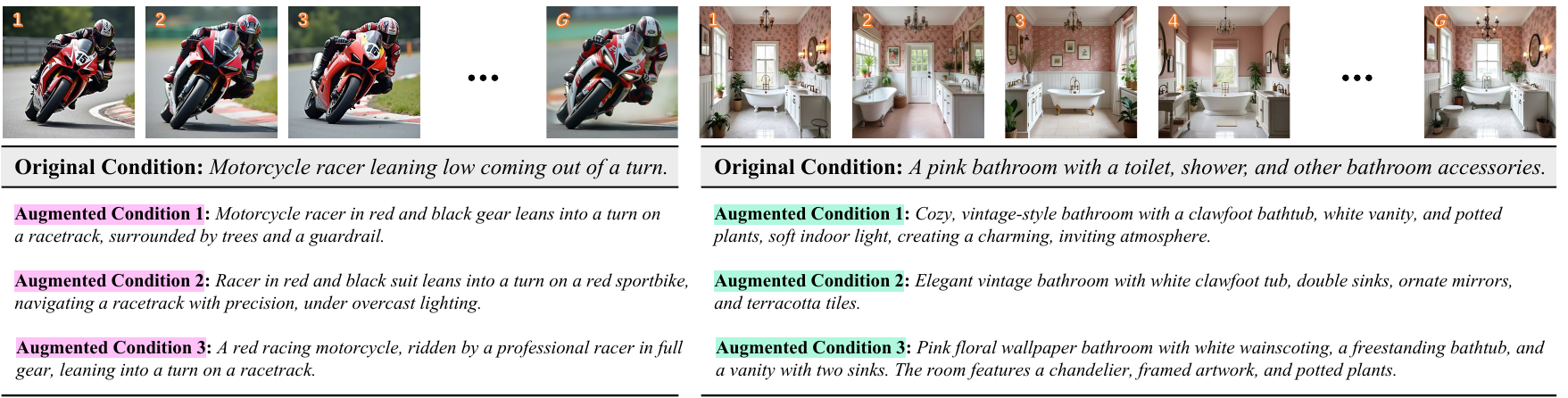}
    \vspace{-2em}
    \caption{
        \textbf{Augmented Conditions.} MV-GRPO generates diverse augmented conditions by leveraging variations among SDE samples and multi-view descriptive prompts.
        }
    \vspace{-0.5em}
    \label{fig:cond}
\end{figure}

\noindent\textbf{Effects of Condition Diversity}. 
The diversity of augmented conditions stems from two aspects: 
(i) each augmented condition is derived from a distinct SDE sample, and (ii) a diverse set of VLM prompts $\mathcal{P}_\text{VLM}$ covering various descriptive perspectives is employed to query the VLM. We ablate (i) by generating all conditions from the same ODE sample, and (ii) by removing the multi-perspective prompt set $\mathcal{P}_\text{VLM}$. 
As shown in Tab.~\ref{tab:ablation} (b), removing either component leads to a notable decline in performance. 
This confirms that both sample-level stochasticity and prompt-level semantic variety are crucial for constructing a robust and diverse augmented condition space. Examples of the augmented conditions used during training are illustrated in Fig.~\ref{fig:cond}.

\begin{table}[t]
\centering
\caption{Ablation experiments on MV-GRPO components and hyperparameters.}
\vspace{-1em}
\label{tab:ablation}
\resizebox{\linewidth}{!}{ 
\begin{tabular}{lccccccccc} 
\toprule
\textbf{Components} & \textbf{Values \& Choices} & {\textbf{HPS-v3}} & {\textbf{UR-v2-A}} & {\textbf{UR-v2-C}} & {\textbf{UR-v2-S}} & \textbf{HPS-v2} & {\textbf{CLIP}} & \textbf{IR} & \textbf{UR-v1} \\ \midrule
\multirow{4}{*}{{(a) Condition Number $K$}} 
& w/o $\mathcal{V}_K$ & 0.149 & \textbf{3.238} & 3.668 & 3.269 & 0.332 & \textbf{0.378} & 1.152 & 3.553 \\
 & $K=3$ & 0.152 & \underline{3.235} & 3.684 & 3.293 & 0.335 & \underline{0.377} & 1.170 & 3.567 \\
 & $K=6$ & \underline{0.154} & 3.233 & \underline{3.696} & \textbf{3.326} & \underline{0.338} & 0.372 & \underline{1.178} & \textbf{3.573} \\
 & \cellcolor{color3}{\textbf{$\mathbf{K=12}$}} & \cellcolor{color3}{\textbf{0.155}} & \cellcolor{color3}{3.229} & \cellcolor{color3}{\textbf{3.701}} & \cellcolor{color3}{\underline{3.320}} & \cellcolor{color3}{\textbf{0.340}} & \cellcolor{color3}{0.370} & \cellcolor{color3}{\textbf{1.193}} & \cellcolor{color3}{\underline{3.569}} \\
 \midrule
\multirow{3}{*}{{(b) Condition Diversity}} 
& w/o SDE query & \underline{0.153} & \underline{3.233} & \underline{3.692} & \underline{3.304} & \underline{0.336} & \underline{0.374} & \underline{1.183} & 3.560 \\
& w/o $\mathcal{P}_\text{VLM}$ & 0.152 & \textbf{3.237} & 3.687 & 3.295 & 0.334 & \textbf{0.376} & 1.174 & \underline{3.564} \\
 & \cellcolor{color3}{\textbf{MV-GRPO}} & \cellcolor{color3}{\textbf{0.155}} & \cellcolor{color3}{3.229} & \cellcolor{color3}{\textbf{3.701}} & \cellcolor{color3}{\textbf{3.320}} & \cellcolor{color3}{\textbf{0.340}} & \cellcolor{color3}{0.370} & \cellcolor{color3}{\textbf{1.193}} & \cellcolor{color3}{\textbf{3.569}} \\
 \midrule
\multirow{2}{*}{{(c) Enhancer Scale}} 
& Qwen3-VL-2B & \underline{0.153} & \textbf{3.238} & \underline{3.694} & \underline{3.317} & \textbf{0.344} & \textbf{0.378} & \underline{1.175} & \textbf{3.575} \\
 & \cellcolor{color3}{\textbf{Qwen3-VL-8B}} & \cellcolor{color3}{\textbf{0.155}} & \cellcolor{color3}{\underline{3.229}} & \cellcolor{color3}{\textbf{3.701}} & \cellcolor{color3}{\textbf{3.320}} & \cellcolor{color3}{\underline{0.340}} & \cellcolor{color3}{\underline{0.370}} & \cellcolor{color3}{\textbf{1.193}} & \cellcolor{color3}{\underline{3.569}} \\
 \bottomrule
\end{tabular}
}
\vspace{-2em}
\end{table}

\noindent\textbf{Effects of Enhancer Scale}. We investigate the impact of the Condition Enhancer's parameter scale by comparing the originally adopted Qwen3-VL-8B with its lightweight variant, Qwen3-VL-2B. As depicted in Tab.~\ref{tab:ablation} (c), Qwen3-VL-8B yields better overall performance on primary metrics such as HPS-v3, IR, and UR-v2-C/S. However, it is noteworthy that the 2B model delivers remarkably competitive results, even marginally surpassing its 8B variant on metrics like UR-v1 and HPS-v2. This indicates that the core advantage of MV-GRPO stems fundamentally from the dense multi-view evaluation mechanism itself, rather than merely relying on the capacity of the Condition Enhancer.

\section{Conclusion}

In this work, we identify that standard flow-based GRPO relies on a sparse, single-view reward evaluation scheme that causes insufficient exploration of intra-group relationships and suboptimal performance. To this end, we introduce MV-GRPO, a novel reinforcement learning framework that shifts the alignment paradigm from single-view to dense, multi-view supervision. MV-GRPO leverages a flexible Condition Enhancer module to augment the condition space with semantically adjacent yet diverse descriptors, enabling a dense multi-view reward mapping that captures rich semantic attributes and provides comprehensive advantage estimation, without the overhead of sample regeneration. Experiments demonstrate MV-GRPO's superiority over existing state-of-the-art methods.

\newpage
\bibliographystyle{splncs04}
\bibliography{main}

\newpage
\newcommand{\appendixhomecontents}{
  \begin{center}
    {\Large\bfseries From Sparse to Dense: Multi-View GRPO for Flow Models via Augmented Condition Space\\
    Supplementary Material\par}
  \end{center}
  \vspace{1em}
}

\appendixhomecontents

\appendix

\section{Overview}

In the supplementary material, we present additional implementation details (Section~\ref{sec:detail}), additional qualitative results (Section~\ref{sec:visual}), all text prompts used in image generation (Section~\ref{sec:prompts}), more discussion on condition enhancement (Section~\ref{sec:discussion}), the limitations of our method (Section~\ref{sec:limitation}), the ethical statement (Section~\ref{sec:ethical}), the reproducibility statement (Section~\ref{sec:reproducibility}), as well as the declaration on LLM usage (Section~\ref{sec:llm}), as a supplement to the main paper. 

\section{Additional Implementation Details}\label{sec:detail}

\subsection{Hyperparameter Configuration}

Tab.~\ref{tab:hyperparams} lists the specific hyperparameter settings employed in our study. These parameters were maintained consistently across all our experiments.
\begin{table}[ht]
\vspace{-1em}
\caption{Hyperparameter settings in our experiments.}
\vspace{-1em}
\centering
\resizebox{0.6\linewidth}{!}{
\begin{tabular}{lclc}
\toprule
\textbf{Parameter} & \textbf{Value} & \textbf{Parameter} & \textbf{Value} \\
\midrule
Random seed & 42 & Learning rate & $2\times 10^{-6}$ \\
Train batch size & 1 & Weight decay & $1\times 10^{-4}$ \\
Warmup steps & 0 & Mixed precision & bfloat16 \\
Dataloader workers & 4 & Max grad norm & 1.0 \\
Eta & 0.7 & Sampler seed & 1223627 \\
Group size & 12 & Scheduler shift  & 3 \\
Sampling steps & 16  & Adv. clip max & 5.0 \\
Init same noise & Yes & Condition Number $K$ & $12$ \\
The number of GPUs & 16 &  Clip range  & $1\times 10^{-4}$\\
\bottomrule
\label{tab:hyperparams}
\end{tabular}
}
\vspace{-3em}
\end{table}

\subsection{Prompts for VLM and LLM Condition Enhancer}

\begin{figure}[h]
    \centering
    \includegraphics[width=1.0\linewidth]{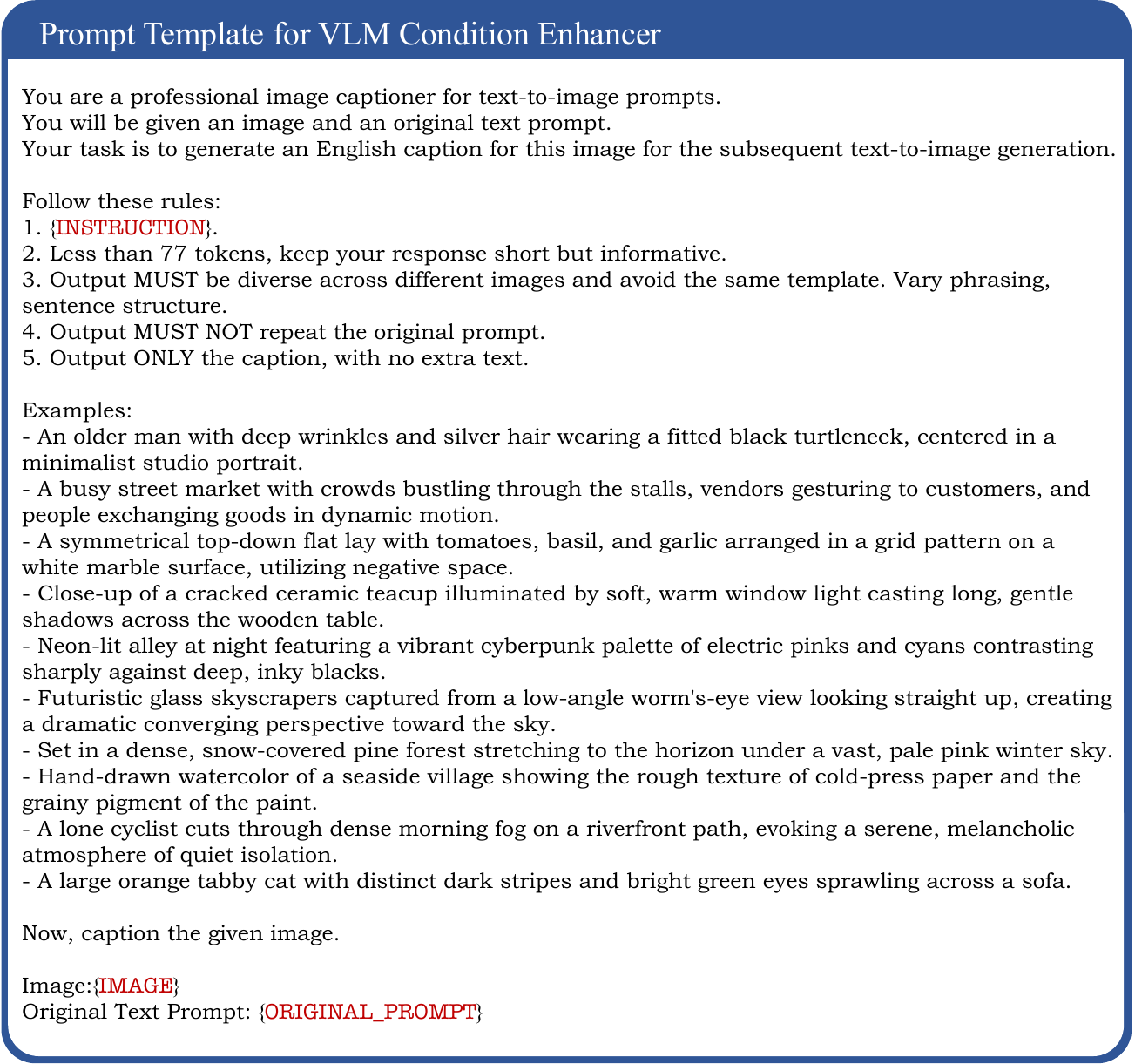}
    \caption{
        \textbf{VLM Prompt Template}. It integrates a descriptive instruction from $\mathcal{P}_\text{VLM}$, along with an image and its original text prompt, to construct a complete VLM prompt for querying the VLM Condition Enhancer to obtain augmented conditions.
        }
    \label{fig:vlm_prompt}
\end{figure}

\textbf{VLM Prompts}. The VLM prompt for MV-GRPO consists of two components: an instruction set $\mathcal{P}_\text{VLM}$ containing diverse descriptive perspectives, and a prompt template. During each VLM query, a specific instruction \texttt{P}$_\text{VLM}$ is randomly sampled from $\mathcal{P}_\text{VLM}$ and inserted into the template to interact with the VLM Condition Enhancer. The content of $\mathcal{P}_\text{VLM}$ is presented in Tab.~\ref{tab:vlm_prompt_set}, while the VLM prompt template is illustrated in Fig.~\ref{fig:vlm_prompt}.

\begin{table}[t]
    \centering
    \caption{List of instructions in the set $\mathcal{P}_\text{VLM}$ used to query the VLM. Each instruction guides the model to focus on a specific visual dimension when generating conditions.}
    \label{tab:vlm_prompt_set}
    \resizebox{0.6\textwidth}{!}{
        \begin{tabular}{c c}
            \toprule
            \textbf{ID} & \textbf{Instructions} \\
            \midrule
            1 & Focus on the main subjects and key attributes. \\
            \rowcolor{gray!10} 2 & Focus on actions, motion, or interactions. \\
            3 & Focus on scene layout and spatial relationships. \\
            \rowcolor{gray!10} 4 & Focus on lighting conditions and shadows. \\
            5 & Focus on color palette and contrast. \\
            \rowcolor{gray!10} 6 & Focus on camera angle, lens feel, and depth of field. \\
            7 & Focus on background/setting and environmental context. \\
            \rowcolor{gray!10} 8 & Focus on textures and material details. \\
            9 & Focus on mood, atmosphere, and tone. \\
            \bottomrule
        \end{tabular}
    }
\end{table}

\clearpage
\newpage

\begin{figure}[h]
    \centering
    \includegraphics[width=1.0\linewidth]{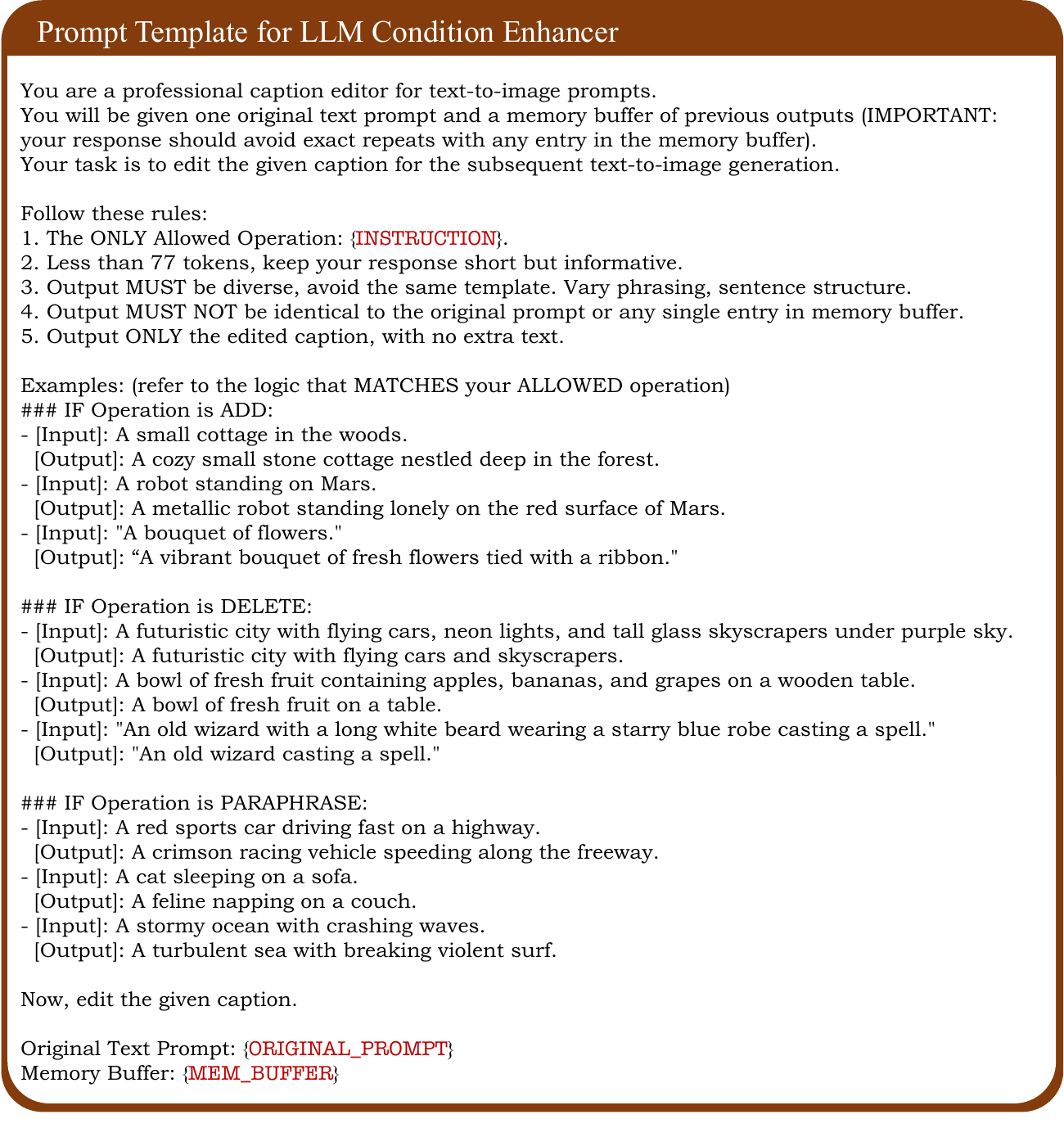}
    \caption{
        \textbf{LLM Prompt Template}. This template incorporates a specific operation from $\mathcal{P}_\text{LLM}$
  and a memory buffer of prior outputs, guiding the LLM Condition Enhancer to refine the original text prompt into diverse augmented conditions.
        }
    \label{fig:llm_prompt}
\end{figure}

\noindent \textbf{LLM Prompts}. Similar to the VLM enhancer setting, the LLM prompt for MV-GRPO also features two components: an instruction set $\mathcal{P}_\text{LLM}$ containing three operations (\texttt{addition}, \texttt{deletion}, and \texttt{rewriting}), and a prompt template.
For each LLM query, an operation $\texttt{P}_\text{LLM}$ is randomly selected from $\mathcal{P}_\text{LLM}$ and incorporated into the template to facilitate interaction with the LLM Condition Enhancer.
The details of $\mathcal{P}_\text{LLM}$ are listed in Tab.~\ref{tab:llm_prompt_set}, and the LLM prompt template is depicted in Fig.~\ref{fig:llm_prompt}.

\begin{table}[t]
    \centering
    \caption{List of operations in the set $\mathcal{P}_\text{LLM}$ used to query the LLM. Each operation directs the model to modify the input prompt while maintaining semantic consistency.}
    \label{tab:llm_prompt_set}
    \resizebox{0.75\textwidth}{!}{
        \begin{tabular}{c >{\centering\arraybackslash}p{11cm}}
            \toprule
            \textbf{ID} & \textbf{Instructions} \\
            \midrule
            1 & ADD: Expand the prompt by enriching existing elements with descriptive details (<= 20 tokens), strictly avoiding the introduction of new subjects. \\
            \rowcolor{gray!10} 2 & DELETE: Remove some non-essential details (<= 20 tokens) to produce a concise caption, but the output must remain a valid description of image.  \\
            3 & PARAPHRASE: Swap words/phrases for close synonyms (adjectives -> near-synonyms; nouns -> another common term for the same thing). \\
            \bottomrule
        \end{tabular}
    }
\end{table} 

\section{Additional Qualitative Results}\label{sec:visual}

\begin{figure}[h]
    \centering
    \includegraphics[width=1.0\linewidth]{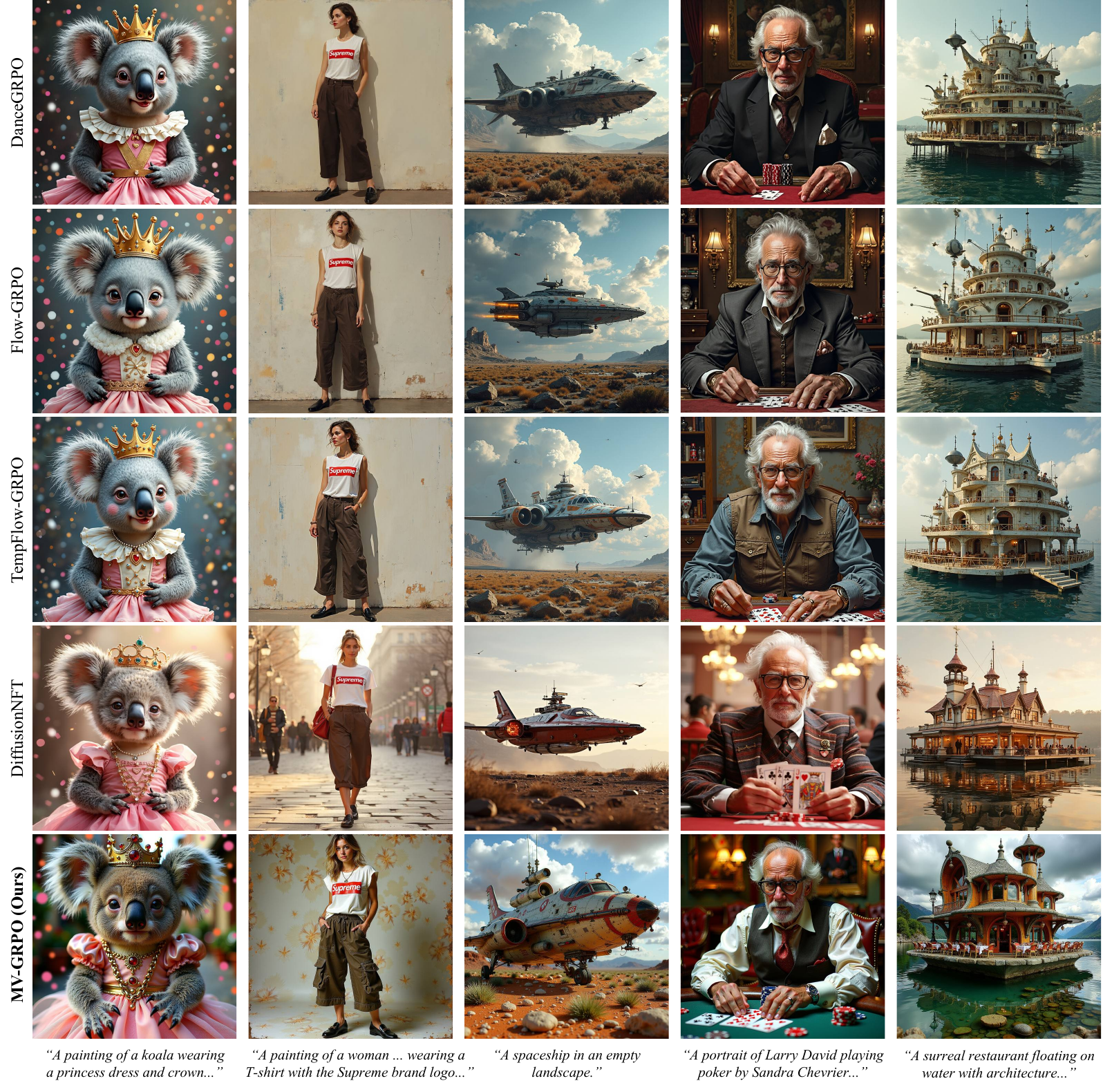}
    \vspace{-1em}
    \caption{
        \textbf{Additional Comparison Results on HPS-v3. (1/2)}
        }
    \label{fig:hpsv3-1}
\end{figure}

\begin{figure}[h]
    \centering
    \includegraphics[width=1.0\linewidth]{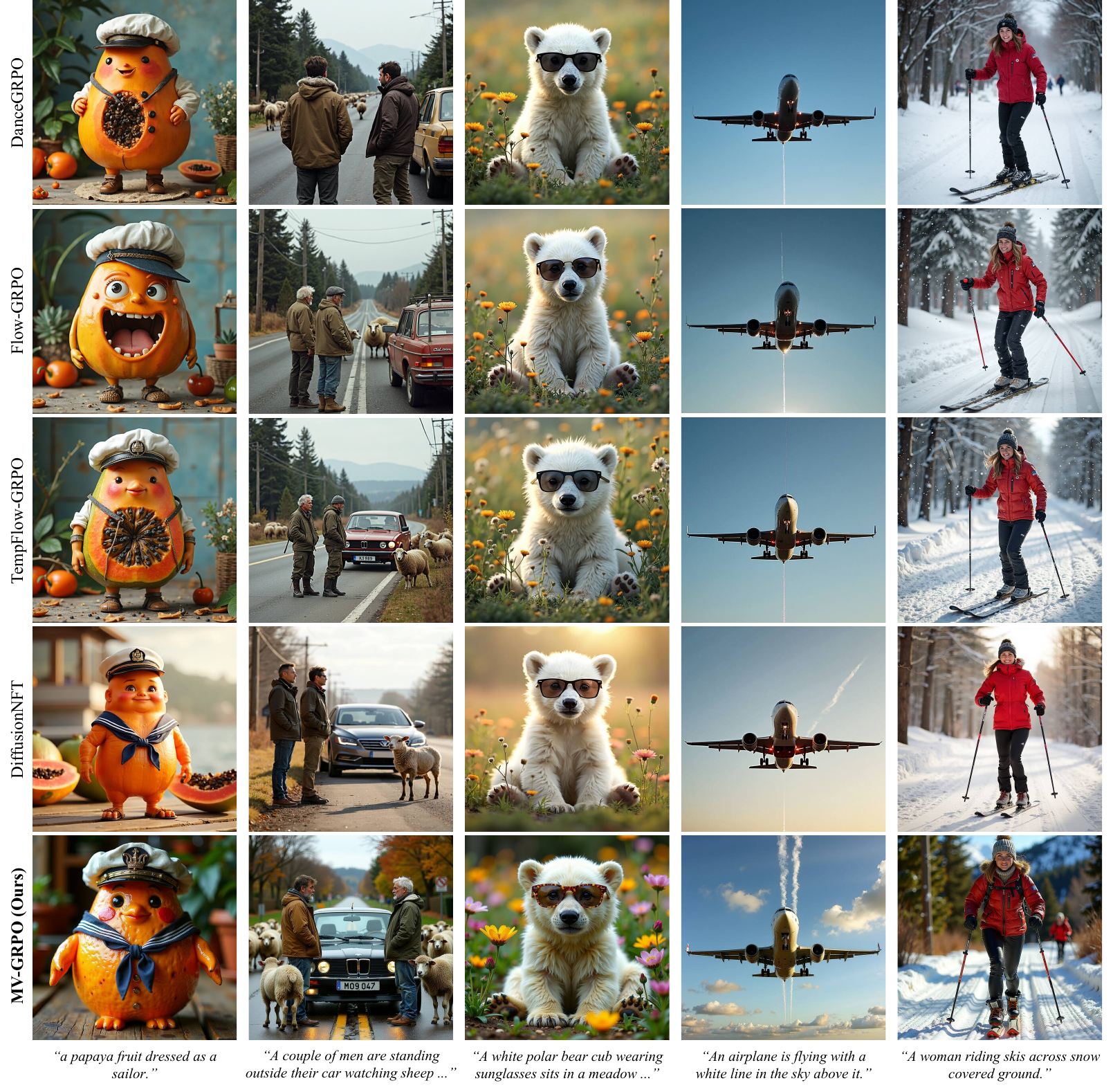}
    \vspace{-1em}
    \caption{
        \textbf{Additional Comparison Results on HPS-v3. (2/2)}
        }
    \label{fig:hpsv3-2}
\end{figure}

\begin{figure}[h]
    \centering
    \includegraphics[width=1.0\linewidth]{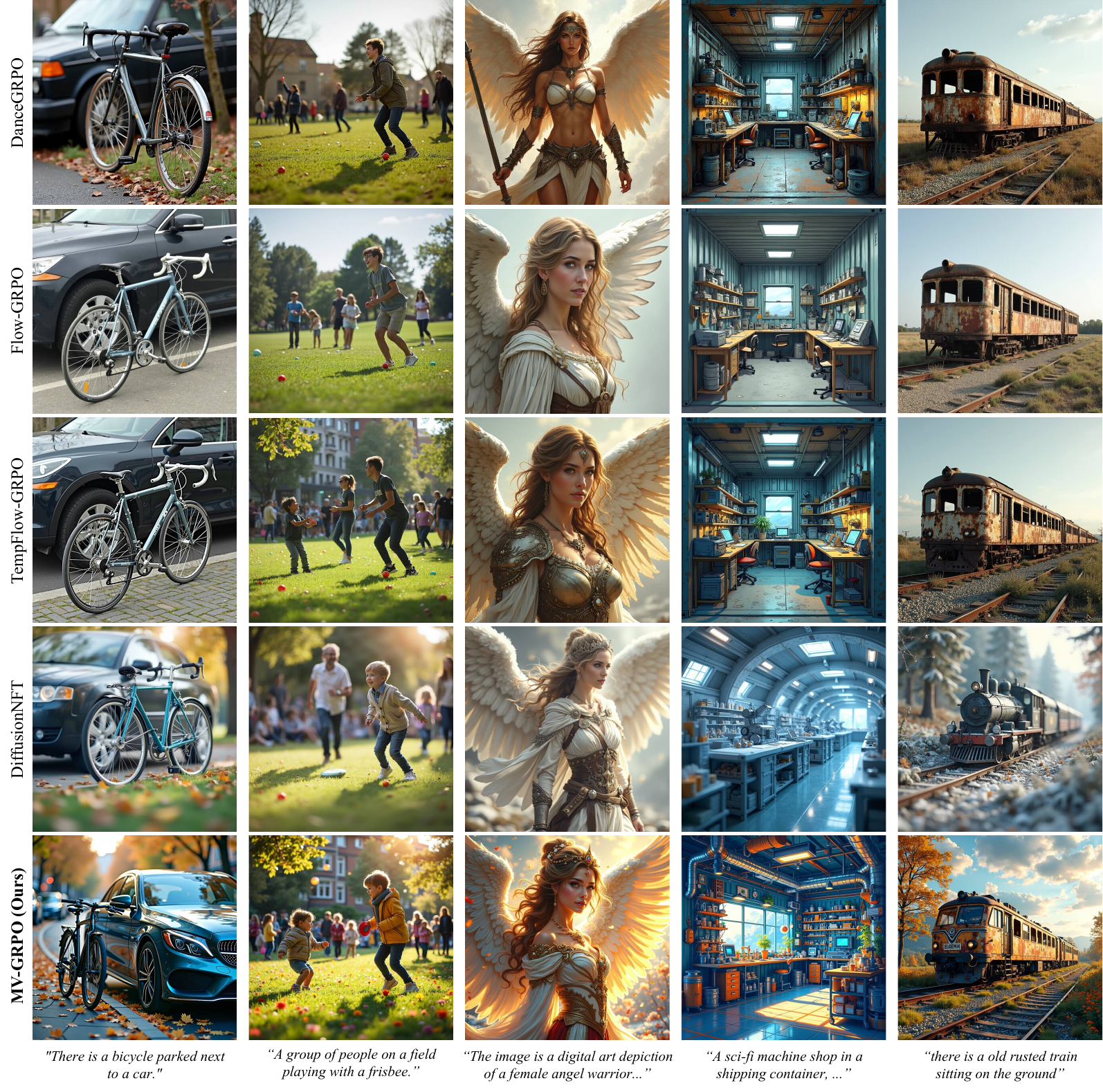}
    \vspace{-1em}
    \caption{
        \textbf{Additional Comparison Results on UnifiedReward-v2. (1/2)}
        }
    \label{fig:ur-1}
\end{figure}

\begin{figure}[h]
    \centering
    \includegraphics[width=1.0\linewidth]{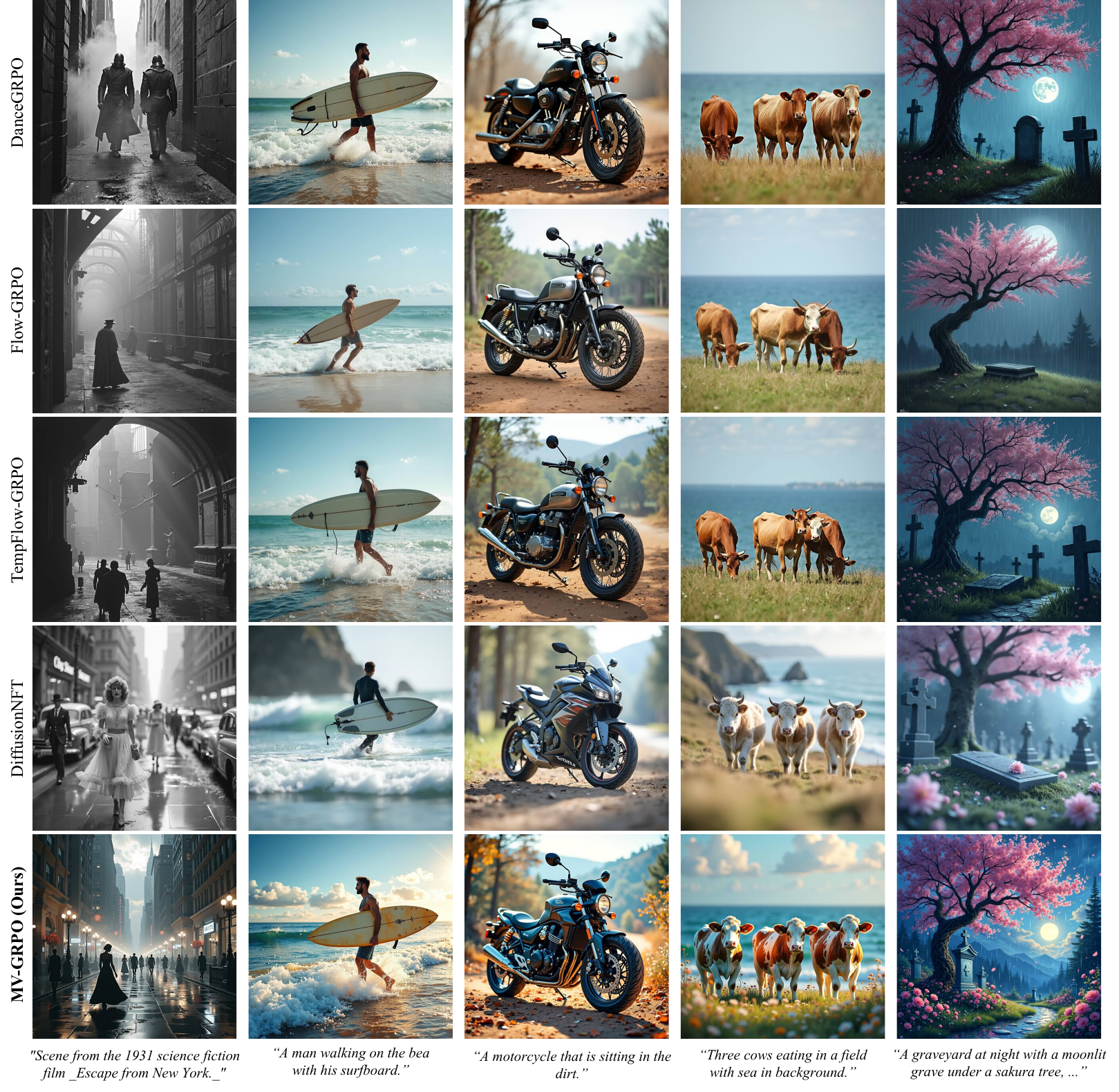}
    \vspace{-1em}
    \caption{
        \textbf{Additional Comparison Results on UnifiedReward-v2. (2/2)}
        }
    \label{fig:ur-2}
\end{figure}

\begin{figure}[h]
    \centering
    \includegraphics[width=1.0\linewidth]{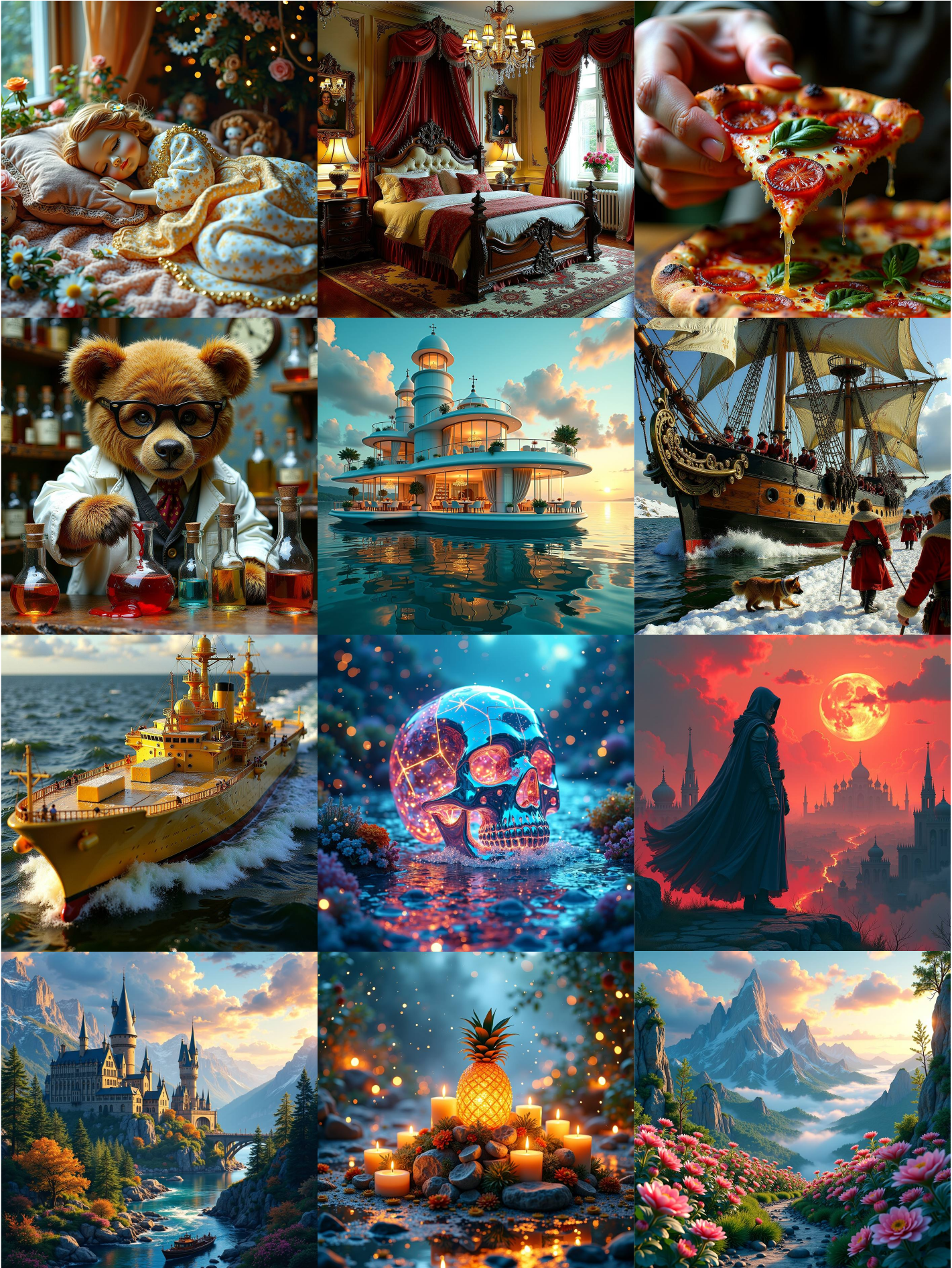}
    \vspace{-1em}
    \caption{
        \textbf{Additional Visual Samples of MV-GRPO. (1/4)}
        }
    \label{fig:gallery-1}
\end{figure}

\begin{figure}[h]
    \centering
    \includegraphics[width=1.0\linewidth]{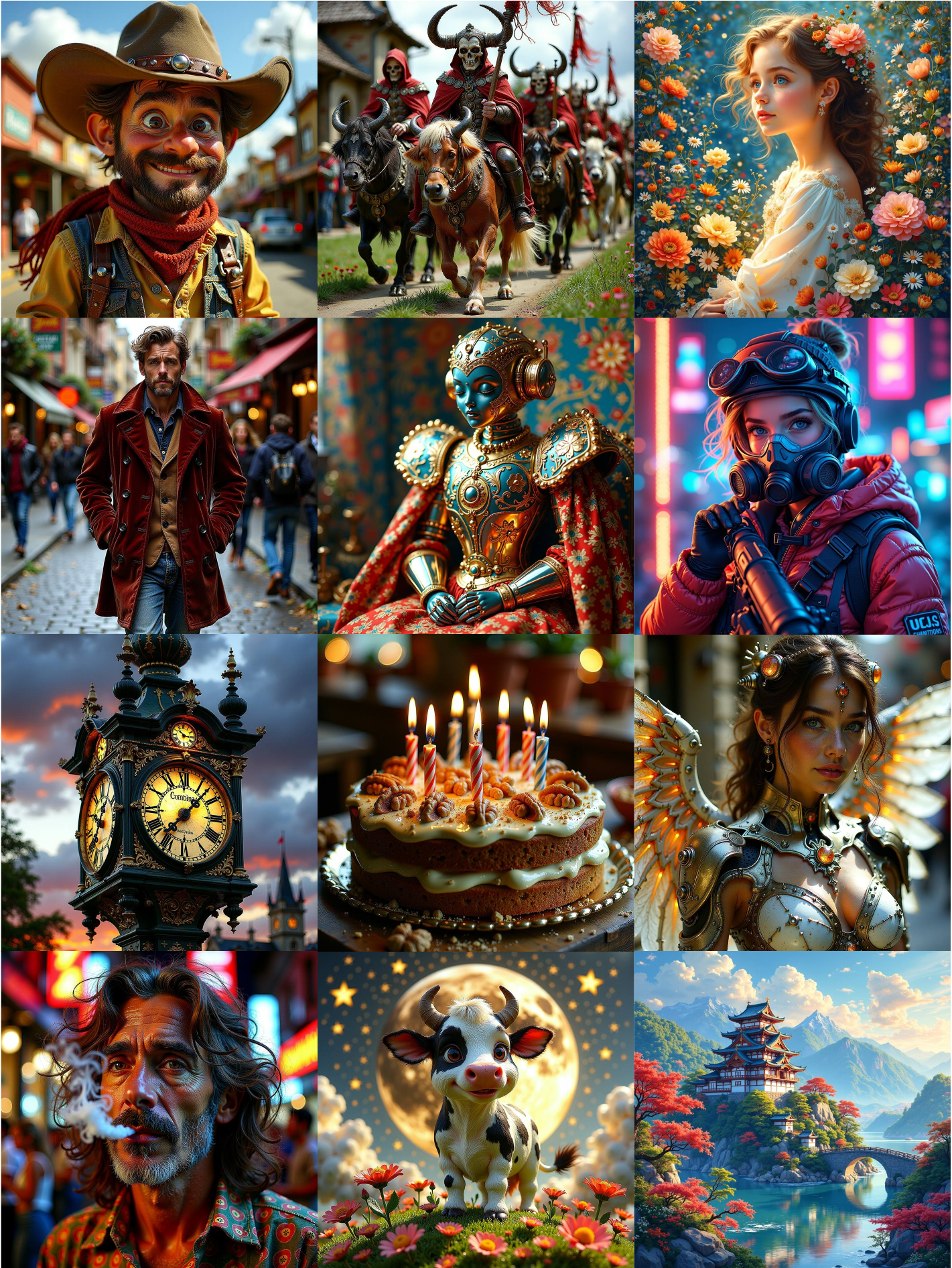}
    \vspace{-1em}
    \caption{
        \textbf{Additional Visual Samples of MV-GRPO. (2/4)}
        }
    \label{fig:gallery-2}
\end{figure}

\begin{figure}[h]
    \centering
    \includegraphics[width=1.0\linewidth]{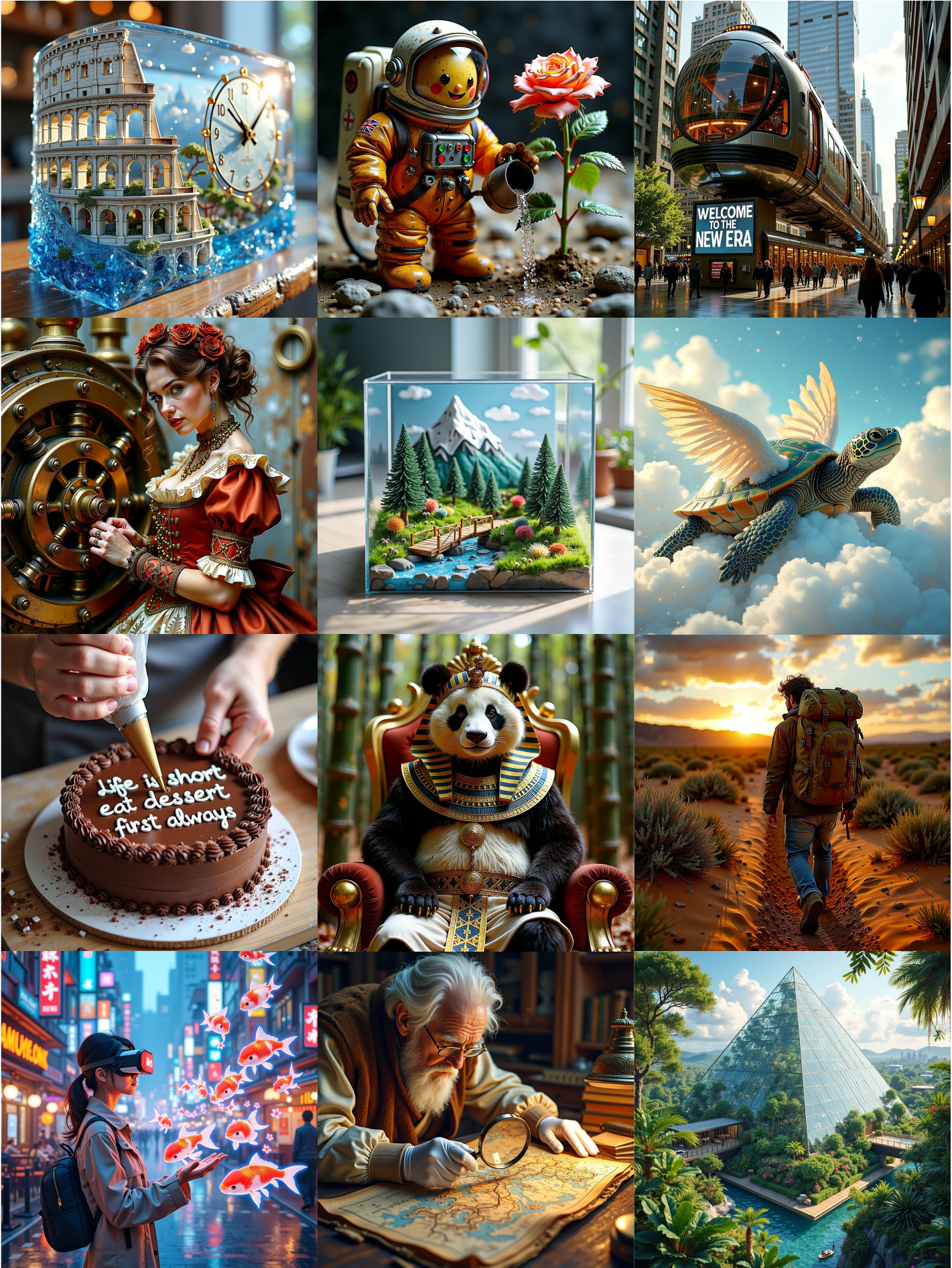}
    \vspace{-1em}
    \caption{
        \textbf{Additional Visual Samples of MV-GRPO. (3/4)}
        }
    \label{fig:gallery-4}
\end{figure}

\begin{figure}[h]
    \centering
    \includegraphics[width=1.0\linewidth]{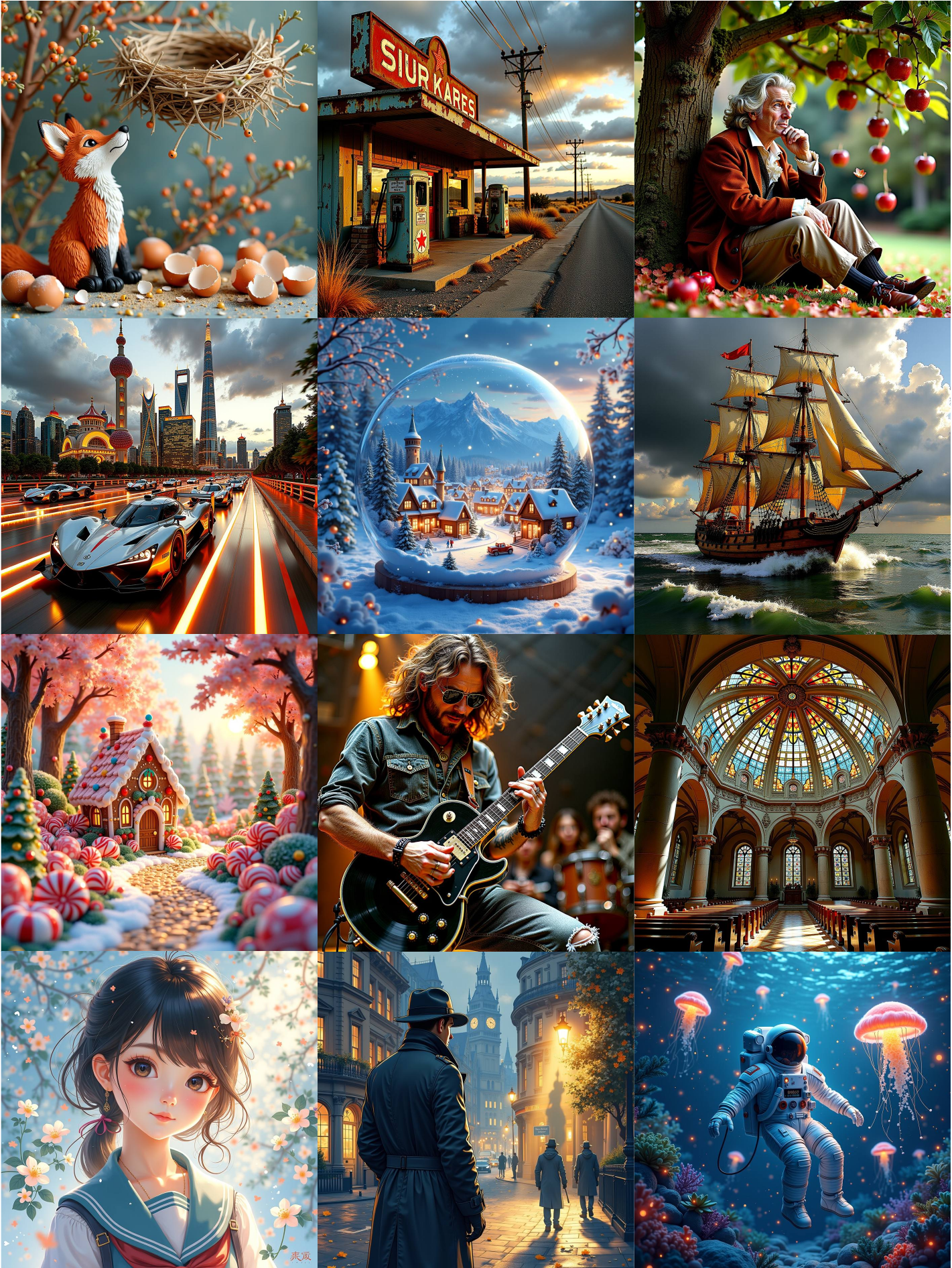}
    \vspace{-1em}
    \caption{
        \textbf{Additional Visual Samples of MV-GRPO. (4/4)}
        }
    \label{fig:gallery-5}
\end{figure}

\begin{figure}[h]
    \centering
    \includegraphics[width=0.9\linewidth]{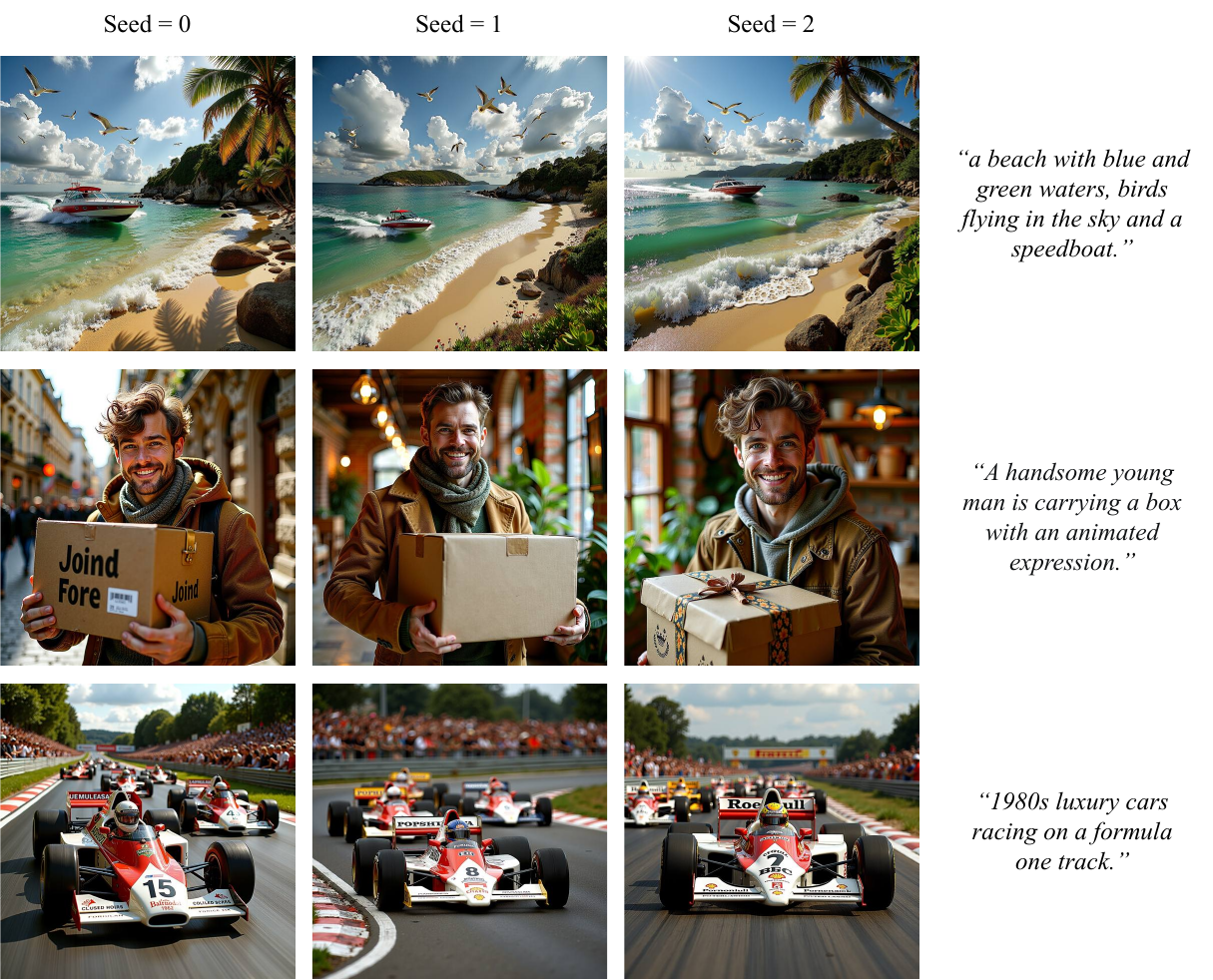}
    \vspace{-0.5em}
    \caption{
        \textbf{Results using same prompts and different seeds. (HPS-v3)}
        }
    \label{fig:seed-hps}
\end{figure}

\begin{figure}[h]
    \centering
    \includegraphics[width=0.9\linewidth]{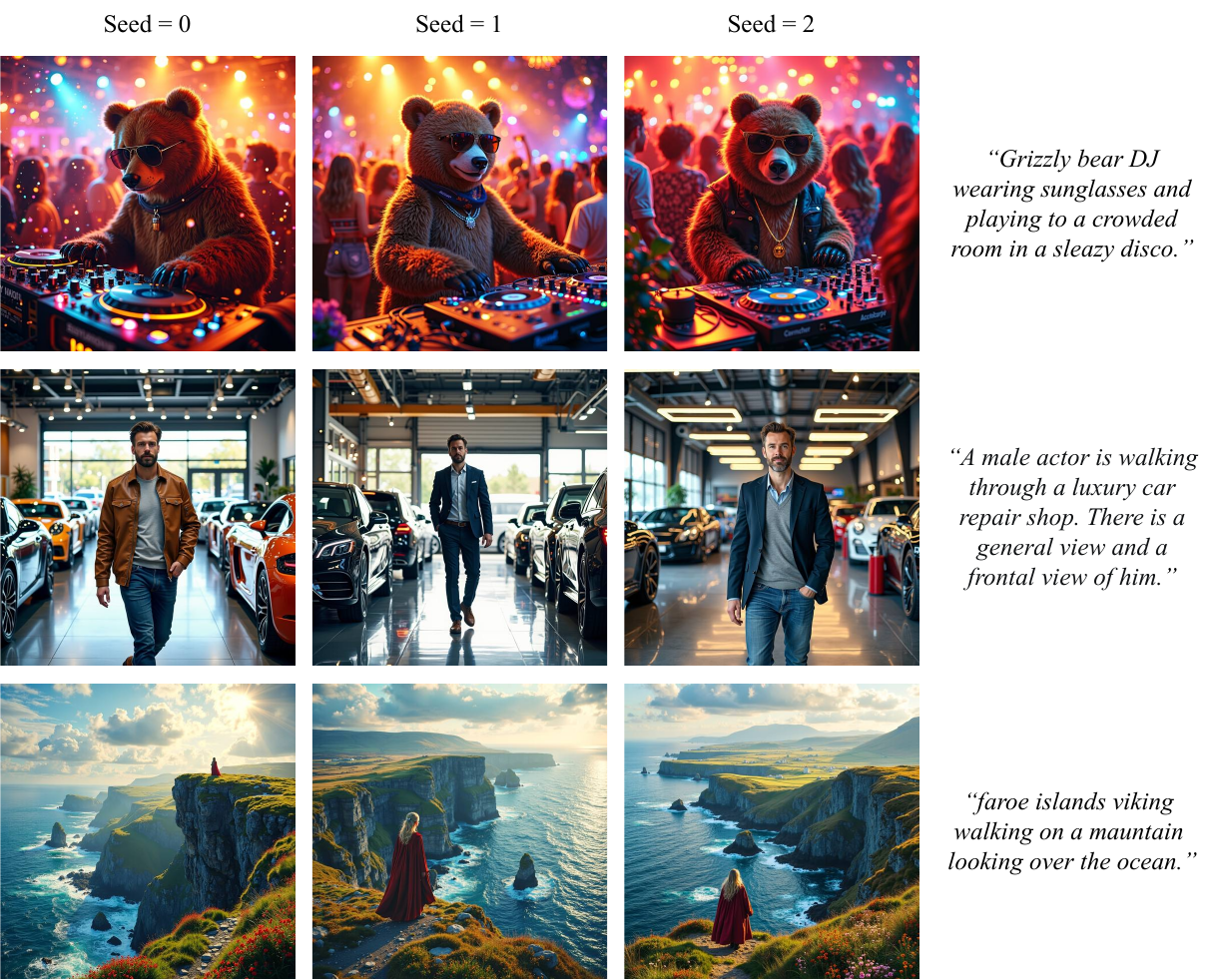}
    \vspace{-0.5em}
    \caption{
        \textbf{Results using same prompts and different seeds. (UR-v2)}
        }
    \label{fig:seed-ur}
\end{figure}

In this section, we present more qualitative comparisons between the proposed MV-GRPO and existing flow-based GRPO methods in Fig.~\ref{fig:hpsv3-1}, Fig.~\ref{fig:hpsv3-2}, Fig.~\ref{fig:ur-1} and Fig.~\ref{fig:ur-2}, additional visual results of MV-GRPO in Fig.~\ref{fig:gallery-1}, Fig.~\ref{fig:gallery-2}, Fig.~\ref{fig:gallery-4} and Fig.~\ref{fig:gallery-5}, along with generated results using the same prompts and different random seeds ($0/1/2$) in Fig.~\ref{fig:seed-hps} and Fig.~\ref{fig:seed-ur}.

\section{Text Prompts for Image Generation}\label{sec:prompts}

All prompts used to generate images in this paper are listed in Tab.~\ref{tab:prompt-1} and Tab.~\ref{tab:prompt-2}.

\section{More Discussion on Condition Enhancement}\label{sec:discussion}

\begin{figure}[t]
    \centering
    \includegraphics[width=0.8\linewidth]{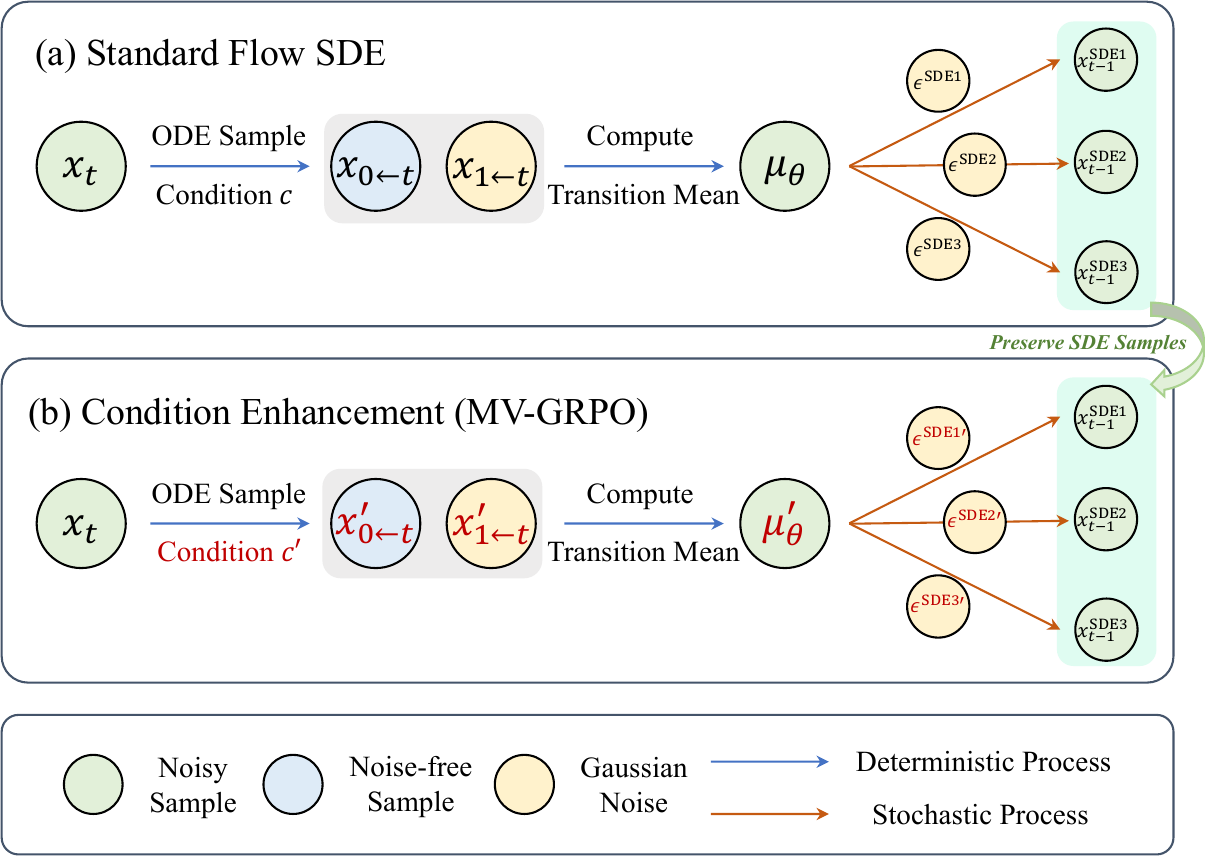}
    \vspace{-0.5em}
    \caption{
    \textbf{Illustration of Equivalent SDE Noise.} 
    (a) Standard flow SDE generates samples based on the transition mean anchored to a single condition $\mathbf{c}$. 
    (b) When substituting $\mathbf{c}$ with an augmented condition $\mathbf{c}'$, the transition mean shifts accordingly. To reach the original samples, an equivalent noise term $\boldsymbol{\epsilon}^{\text{SDE}'}$ is implicitly required.
    The $\boldsymbol{\mu}_\theta$ and $\boldsymbol{\mu}'_\theta$ in the figure are shorthand for $\boldsymbol{\mu}_\theta(\boldsymbol{x}_t, \mathbf{c})$ and $\boldsymbol{\mu}_\theta(\boldsymbol{x}_t, \mathbf{c}')$, respectively. 
}
\vspace{-1.5em}
    \label{fig:discussion}
\end{figure}

In this section, we provide a deeper analysis on condition enhancement for flow-based GRPO. Specifically, we rationalize the validity of using augmented conditions to optimize original samples through the lens of \textit{Equivalent SDE Noise}. 

\noindent \textbf{Standard Flow SDE}. First, we revisit the standard flow SDE widely adopted in existing GRPO frameworks \cite{liu2025flow, xue2025dancegrpo}. As illustrated in Fig.~\ref{fig:discussion} (a), given a noisy sample $\boldsymbol{x}_t$ and a condition $\mathbf{c}$, the flow model first executes a deterministic ODE sampling to estimate the underlying noise-free sample $\boldsymbol{x}_{0\leftarrow t}$ and the corresponding Gaussian noise $\boldsymbol{x}_{1\leftarrow t}$ at the current timestep $t$:
\begin{equation}
    \boldsymbol{x}_{0\leftarrow t} = \boldsymbol{x}_t - t \cdot \boldsymbol{v}_\theta(\boldsymbol{x}_t,t,\mathbf{c}), 
\end{equation}
\begin{equation}
    \boldsymbol{x}_{1\leftarrow t} = \boldsymbol{x}_t + (1-t)\cdot\boldsymbol{v}_\theta(\boldsymbol{x}_t,t,\mathbf{c}),
\end{equation}
where $\boldsymbol{v}_\theta(\boldsymbol{x}_t,t,\mathbf{c})$ denotes the predicted flow velocity given condition $\mathbf{c}$. Subsequently, $\boldsymbol{x}_{0\leftarrow t}$ and $\boldsymbol{x}_{1\leftarrow t}$ are combined to yield the transition mean $\boldsymbol{\mu}_\theta(\boldsymbol{x}_t, \mathbf{c})$ of the current SDE rollout: 
\begin{equation}
    \boldsymbol{\mu}_\theta(\boldsymbol{x}_t, \mathbf{c}) = (1 - t - \Delta t)\boldsymbol{x}_{0\leftarrow t} + \left( t + \Delta t + \frac{\sigma_t^2 \Delta t}{2t} \right) \boldsymbol{x}_{1\leftarrow t},
\end{equation}
in which $\Delta t$ stands for the step size. Finally, each SDE sample $\boldsymbol{x}_{t-1}^\text{SDE}$ is obtained by injecting a randomly sampled Gaussian noise $\boldsymbol{\epsilon}^\text{SDE}$ into $\boldsymbol{\mu}_\theta(\boldsymbol{x}_t, \mathbf{c})$:
\begin{equation}
    \boldsymbol{x}_{t-1}^\text{SDE} = \boldsymbol{\mu}_\theta(\boldsymbol{x}_t, \mathbf{c}) + \sigma_t \sqrt{\Delta t} \boldsymbol{\epsilon}^\text{SDE}.
\end{equation}
In Fig.~\ref{fig:discussion} (a), we denote the specific Gaussian noise components sampled during independent SDE rollouts as $\boldsymbol{\epsilon}^\text{SDE1}$, $\boldsymbol{\epsilon}^\text{SDE2}$ and $\boldsymbol{\epsilon}^\text{SDE3}$, which consequently lead to distinct SDE samples $\boldsymbol{x}_{t-1}^\text{SDE1}$, $\boldsymbol{x}_{t-1}^\text{SDE2}$ and $\boldsymbol{x}_{t-1}^\text{SDE3}$, respectively. 

\noindent\textbf{Condition Enhancement}. As illustrated in Fig.~\ref{fig:discussion} (b), the Condition Enhancement of MV-GRPO initiates the process from the identical noisy latent state $\boldsymbol{x}_t$ but substitutes the input condition for the flow model with an augmented view \textcolor{blue}{$\mathbf{c}'$}. 
Given the shift in flow velocity (from $\boldsymbol{v}_\theta(\boldsymbol{x}_t,t,\mathbf{c})$ to $\boldsymbol{v}_\theta(\boldsymbol{x}_t,t,\textcolor{blue}{\mathbf{c}'})$), the deterministic estimates for the clean sample and the Gaussian noise will both undergo corresponding transformations:
\begin{equation}
    \boldsymbol{x}_{0\leftarrow t} \rightarrow \boldsymbol{x}_{0\leftarrow t}^{'}, \quad \boldsymbol{x}_{1\leftarrow t} \rightarrow \boldsymbol{x}_{1\leftarrow t}^{'}.
\end{equation}
Therefore, the transition mean also changes accordingly:
\begin{equation}
    \boldsymbol{\mu}_\theta(\boldsymbol{x}_t, \mathbf{c}') = (1 - t - \Delta t)\boldsymbol{x}_{0\leftarrow t}^{'} + \left( t + \Delta t + \frac{\sigma_t^2 \Delta t}{2t} \right) \boldsymbol{x}_{1\leftarrow t}^{'}.
\end{equation}
If we expect the updated mean $\boldsymbol{\mu}_\theta(\boldsymbol{x}_t, \mathbf{c}')$ to still transition to the original SDE state $\boldsymbol{x}_{t-1}^\text{SDE}$,
an equivalent noise term $\boldsymbol{\epsilon}^\text{SDE}{'}$ satisfying the following relationship should be explicitly sampled:
\begin{equation}
    \boldsymbol{x}_{t-1}^\text{SDE} = \boldsymbol{\mu}_\theta(\boldsymbol{x}_t, \mathbf{c}') + \sigma_t \sqrt{\Delta t} \textcolor{blue}{\boldsymbol{\epsilon}^\text{SDE}{'}}.
\end{equation}
Therefore, utilizing the augmented condition $\mathbf{c}'$ to optimize the original SDE sample $\boldsymbol{x}_{t-1}^\text{SDE}$ fundamentally hinges on the proximity between the probability of sampling the original noise $\boldsymbol{\epsilon}^\text{SDE}$ and that of the equivalent noise $\boldsymbol{\epsilon}^\text{SDE}{'}$. 

Since measuring the sampling probability discrepancy between stochastic noise terms is mathematically equivalent to quantifying the divergence between the two probability densities $p_\theta(\boldsymbol{x}_{t-1} | \boldsymbol{x}_t, \mathbf{c})$ and $p_\theta(\boldsymbol{x}_{t-1} | \boldsymbol{x}_t, \mathbf{c}')$ (as we discussed in the main paper), this probability gap corresponds directly to the probability drift $\boldsymbol{\delta}(\mathbf{c}, \mathbf{c}')$ defined in Eq.~\ref{eq:semantic_drift} of the main text, which remains minimal for the vast majority of samples (see Fig.~\ref{fig:distribution} in the main paper).
This can be interpreted as follows: given the semantic similarity between $\mathbf{c}$ and $\mathbf{c}'$, the discrepancy between their induced transition means $\boldsymbol{\mu}_\theta(\boldsymbol{x}_t, \mathbf{c})$ and $\boldsymbol{\mu}_\theta(\boldsymbol{x}_t, \mathbf{c}')$ is sufficiently small. Consequently, the SDE noise required to reach the same target state $\boldsymbol{x}_{t-1}^\text{SDE}$ remains proximate, resulting in a minimal difference in sampling probability.

\section{Limitation}\label{sec:limitation}

Despite the advancements of MV-GRPO in preference alignment for flow models, it faces certain constraints.
First, its effectiveness may be limited in tasks with rigid or predefined conditioning signals (e.g., class-conditional generation on a specific dataset), where meaningful condition enhancements are difficult to formulate.
Furthermore, the quality of augmented conditions is inherently bounded by the visual understanding and reasoning capabilities of current VLMs or LLMs. 
However, we anticipate that this limitation will be naturally mitigated as the performance of these models continues to advance.

\section{Ethical Statement}\label{sec:ethical}

We are committed to maintaining the ethical standards and fostering responsible innovation throughout this research. To the best of our knowledge, our study, including the data, methodologies, and applications involved, does not present any ethical concerns. All experiments were conducted in strict accordance with established ethical frameworks, ensuring the integrity, transparency, and reliability of our research, with careful attention to responsible use.

\section{Reproducibility Statement}\label{sec:reproducibility}

To ensure full reproducibility and support the broader research community, we will publicly release the source code of MV-GRPO. 
We envision these resources serving as a valuable baseline for future research in flow-based GRPO, facilitating further innovation and progress in the field.

\section{Declaration on LLM Usage}\label{sec:llm}

In this paper, we use LLMs only for minor language polishing.

\newpage
\vspace*{\fill}
\vspace*{\fill}

\clearpage
\begin{table}[t]
    \centering
    \caption{T2I prompts used in this paper (1/2).  
    Prompts for each figure are listed sequentially, following the order from left to right and top to bottom.}
    \vspace{-1em}
     \label{tab:prompt-1}
     \resizebox{\textwidth}{!}{
        \begin{tabular}{>{\arraybackslash}p{2cm} >{\arraybackslash}p{15cm}}
            \toprule
           \textbf{\centering{Figure}}&\textbf{\centering Text Prompt}\\
            \midrule
            \multirow{16}{*}{Figure.~\ref{fig:teaser}} &{A sleek silver supercar is parked on a wet city street.} \\
 &\cellcolor{color4} A massive and brightly colored spacecraft in a deserted landscape, depicted in retro 1960s sci-fi art. \\    
  & Close-up view of ancient Greek ruins set against a colourful, starry night sky creating a mystical atmosphere.  \\ 
   &\cellcolor{color4} An ornate, lantern-lit courtyard with carved balconies and tiled floors.\\  
   & A beautiful woman in a white gown wears gold jewels, warm outdoor light.  \\
   &\cellcolor{color4} Close-up on a flickering neon installation in a dark futuristic nightclub. The text "Multi-View GRPO" is rendered in intertwining blue and magenta neon tubes, casting colored light onto chrome pipes and exposed wires on a brick wall. Cyberpunk aesthetic, shallow depth of field. \\
   & Two cats, one grey and one black, are wearing steampunk attire and standing in front of a ship in a heavily detailed painting. \\
   &\cellcolor{color4} A bowl filled with apple slices and ice cream. \\
   & A hand-drawn cute gnome holding a pumpkin in an autumn disguise, portrayed in a detailed close-up of the face with warm lighting and high detail. \\
   &\cellcolor{color4} A wooden trunk sitting outside with stickers on it. \\
   & Small carved figurines of fantasy buildings, miniatures and standalone objects on a table. \\
  \midrule
    \multirow{1}{*}{Figure.~\ref{fig:observation}} & A cat and a dog in a teacup.\\
\midrule
    \multirow{5}{*}{Figure.~\ref{fig:hpsv3}} & A blue vase filled with yellow flowers on a window sill.\\
    &\cellcolor{color4} A great room with the living area in the foreground, dining table behind it and kitchen in the very back.  \\
    & A painting of the TV tower by Zaha Hadid. \\
    &\cellcolor{color4} The skater is riding very low on his board. \\
    & A portrait painting of a male deer in a suit sitting on a sofa near a window by John Singer Sargent. \\
\midrule
    \multirow{6}{*}{Figure.~\ref{fig:urv2}} & Male vampire of clan Banu Haqim with blue braided hair stands in a modern city at night surrounded by neon signs, jewelry, and tattoos.\\
    &\cellcolor{color4} A happy daffodil with big eyes, multiple leaf arms and vine legs, rendered in 3D Pixar style.  \\
    & A cave painting of a Stone Age party. \\
    &\cellcolor{color4} A fruit basket on a kitchen table with a Studio Ghibli reference. \\
    & A group of four friends commemorating a ski trip in the snow. \\
\midrule
    \multirow{2}{*}{Figure.~\ref{fig:cond}} & Motorcycle racer leaning low coming out of a turn.\\
    &\cellcolor{color4} A pink bathroom with a toilet, shower, and other bathroom accessories.  \\
\midrule
    \multirow{7}{*}{Figure.~\ref{fig:hpsv3-1}} & A painting of a koala wearing a princess dress and crown, with a confetti background.\\
    &\cellcolor{color4} A painting of a woman by Zinaida Serebriakova wearing a T-shirt with the Supreme brand logo, a sleeveless white blouse, dark brown capris, and black loafers.  \\
    & A spaceship in an empty landscape. \\
    &\cellcolor{color4} A portrait of Larry David playing poker by Sandra Chevrier, featured on Artstation. \\
    & A surreal restaurant floating on water with architecture resembling Salvador Dalí's art, depicted in a film directed by Denis Villeneuve. \\
\midrule
    \multirow{5}{*}{Figure.~\ref{fig:hpsv3-2}} & a papaya fruit dressed as a sailor.\\
    &\cellcolor{color4} A couple of men are standing outside their car watching sheep cross a road. \\
    & A white polar bear cub wearing sunglasses sits in a meadow with flowers. \\
    &\cellcolor{color4} An airplane is flying with a white line in the sky above it. \\
    & A woman riding skis across snow covered ground. \\
\midrule
    \multirow{7}{*}{Figure.~\ref{fig:ur-1}} & There is a bicycle parked next to a car. \\
    &\cellcolor{color4} A group of people on a field playing with a frisbee. \\
    & The image is a digital art depiction of a female angel warrior with detailed features by artist Magali Villeneuve. \\
    &\cellcolor{color4} A sci-fi machine shop in a shipping container, depicted in a manga-style digital painting with intricate details. \\
    & there is a old rusted train sitting on the ground \\
\midrule
    \multirow{5}{*}{Figure.~\ref{fig:ur-2}} & Scene from the 1931 science fiction film \_Escape from New York.\_ \\
    &\cellcolor{color4} A man walking on the bea with his surfboard.  \\
    & A motorcycle that is sitting in the dirt. \\
    &\cellcolor{color4} Three cows eating in a field with sea in background. \\
    & A graveyard at night with a moonlit grave under a sakura tree, rain falling, by Aleksandra Waliszewska. \\
    \midrule
            \multirow{15}{*}{Figure.~\ref{fig:gallery-1}} &A full-body wide-angle photo of a wooden art doll by Agostino Arrivabene, depicting a peaceful sleeping pose. \\
 &\cellcolor{color4} Very ornate bedroom with a chandelier over the bed. \\    
  & A person holding a very small slice on pizza between their fingers.  \\ 
   &\cellcolor{color4} A teddy bear mad scientist mixing chemicals depicted in oil painting style as a fantasy concept art piece.\\  
   & A surreal restaurant floating on water with architecture resembling Salvador Dalí's art, depicted in a film directed by Denis Villeneuve.  \\
   &\cellcolor{color4} A polar expedition unloads from a ship in the 19th century in an intricate and elegant fantasy illustration. \\
   & An image of an aircraft carrier made of cheese. \\
   &\cellcolor{color4} A photograph of a giant diamond skull in the ocean, featuring vibrant colors and detailed textures. \\
   & Hooded figure standing over a ruined city with red haze and a grin. \\
   &\cellcolor{color4} Panorama of Hogwarts. \\
   & An alter made of bones with a glowing pineapple lamp on it and surrounded by candles, in front of a swirling mist with epic lighting. \\
   &\cellcolor{color4} A sunrise-lit mountain range rises above a misty valley, with flowers. \\
\bottomrule
        \end{tabular}
    }
\end{table}
\begin{table}[t]
    \centering
    \caption{T2I prompts used in this paper (2/2). 
    Prompts for each figure are listed sequentially, following the order from left to right and top to bottom.}
    \vspace{-1em}
     \label{tab:prompt-2}
     \resizebox{\textwidth}{!}{
        \begin{tabular}{>{\arraybackslash}p{2cm} >{\arraybackslash}p{15cm}}
            \toprule
           \textbf{\centering{Figure}}&\textbf{\centering Text Prompt}\\
  \midrule
    \multirow{14}{*}{Figure.~\ref{fig:gallery-2}} & American cowboy with a scruffy appearance in a retrofuturistic style, inspired by the animations of Studio Ghibli. \\
 &\cellcolor{color4} Undead army on riding beasts with symbols and music. \\    
  & A Monet portrait. \\ 
   &\cellcolor{color4} a man walking alone down the street in a velvet jacket.\\  
   & a shiny metallic renaissance steampunk robot in the style of Jan van Eyck.  \\
   &\cellcolor{color4} A key visual of a young female swat officer with a neon futuristic gas mask in a cyberpunk setting. \\
   & A clock tower with lighted clock faces, against a twilight sky. \\
   &\cellcolor{color4} A lighted birthday cake with chunks of walnuts. \\
   & A photo of a mechanical angel woman with crystal wings, in the sci-fi style of Stefan Kostic, created by Stanley Lau and Artgerm. \\
   &\cellcolor{color4} A portrait of Frank Zappa smoking, with vivid neon colors, by various artists. \\
   & A cute cartoon cow stands on a grassy hill under a large moon, surrounded by stars and flowers. \\
   &\cellcolor{color4} A Japanese castle landscape painting trending on Artstation. \\
   \midrule
\multirow{20}{*}{Figure.~\ref{fig:gallery-4}} & A crystal wall clock in the shape of an ancient Roman Colosseum, inside the clock is a miniature city. \\
 &\cellcolor{color4} A potato astronaut in a spacesuit is watering a rose plant on the moon. It looks happy, Cyberpunk style. \\    
  & In a huge glass-domed city, a suspended train is silently gliding through the air, and the words ``Welcome to the New Era'' are flashing on a holographic billboard in the center of the city. \\ 
   &\cellcolor{color4} A steampunk picture of a lady in a classic dress operating a complex brass mechanism. \\  
   & A miniature landscape in a glass box, with a small wooden bridge in the foreground, several small trees in the middle, and a model of a snowy mountain in the background.  \\
   &\cellcolor{color4} A turtle with feathered wings is not in the air because it cannot fly, but lies leisurely on the clouds to rest. \\
   & A baker is using a decorative bag to write on a chocolate cake with the English words ``Life is short eat dessert first always'' written on the cake. \\
   &\cellcolor{color4} A giant panda dressed in the costume of an ancient Egyptian pharaoh sits upright on a golden throne in the dense bamboo forest. \\
   & In the golden desert at dusk, a traveler carrying a huge backpack is struggling to walk into the distance, trailing a long shadow behind him. \\
   &\cellcolor{color4} On the street of the future of cyberpunk-style Tokyo, a woman wearing VR glasses controls the holographic koi floating in front of her through the air. \\
   & The texture of the movie. An elderly historian wearing white cotton gloves carefully examined a yellowed sheepskin scroll map with a magnifying glass, with a solemn expression.\\
   &\cellcolor{color4} A huge glass greenhouse shaped like the Great Pyramid of Giza contains a complete and miniature Amazon rainforest ecosystem, and the overall surrealist style.\\
   \midrule
\multirow{21}{*}{Figure.~\ref{fig:gallery-5}} & In a clay sculpture, a hungry fox looks up at an empty bird's nest on a high branch, with broken eggshells scattered on the ground below. \\
&\cellcolor{color4} At dusk, an abandoned gas station stands alone beside the desert road. The wind and sand blowing against the rusty signs, presenting the bleak style of a road movie. \\  
  & The famous scene of physicist Newton sitting under an apple tree, falling into thought after being hit by a falling apple. \\ 
   &\cellcolor{color4} In future Shanghai, the Oriental Pearl Tower is surrounded by a glowing air racing track, and several streamlined racing cars are speeding at high speed. \\  
   & A crystal clear crystal ball contains a quiet town covered with heavy snow. The town is brightly lit and full of a warm winter atmosphere.  \\
   &\cellcolor{color4} A majestic tall wooden ship with sails cuts through ocean waves under a dramatic cloudy sky. \\
   & A forest of huge candies and biscuits with a winding path leading to a mini gingerbread house in a clay animation style. \\
   &\cellcolor{color4} A long-haired rock musician in sunglasses plays a black electric guitar on stage under stage lights. \\
   & Inside an ancient church, the dome is composed of countless stained glass sheets, and the picture presents a solemn cinematic realistic style. \\
   &\cellcolor{color4} A semirealistic digital painting of a Japanese schoolgirl in a gentle grayish color palette, by Chinese artists on ArtStation. \\
   & A detective standing on a street in London, observing a shadow on the opposite building. The picture presents a suspenseful atmosphere.\\
   &\cellcolor{color4} An astronaut floated in the deep sea, surrounded by glowing jellyfish.\\
   \midrule
\multirow{3}{*}{Figure.~\ref{fig:seed-hps}} & a beach with blue and green waters, birds flying in the sky and a speedboat. \\
 &\cellcolor{color4} A handsome young man is carrying a box with an animated expression. \\    
  & 1980s luxury cars racing on a formula one track. \\ 
  \midrule
\multirow{4}{*}{Figure.~\ref{fig:seed-ur}} & Grizzly bear DJ wearing sunglasses and playing to a crowded room in a sleazy disco. \\
 &\cellcolor{color4} A male actor is walking through a luxury car repair shop. There is a general view and a frontal view of him. \\    
  & faroe islands viking walking on a mauntain looking over the ocean. \\ 
\bottomrule
        \end{tabular}
    }
\end{table}
\end{document}